\documentclass{article}
\usepackage{amsmath}
 \usepackage{rotating}
\usepackage{arxiv}
\usepackage{ragged2e}
\usepackage{placeins}
\usepackage{float} 
\usepackage[table]{xcolor} 
\usepackage{dsfont} 
\usepackage[utf8]{inputenc} 
\usepackage[T1]{fontenc}    
\usepackage{hyperref}       
\usepackage{url}            
\usepackage{booktabs}       
\usepackage{amsfonts}       
\usepackage{nicefrac}       
\usepackage{microtype}      
\usepackage{lipsum}
\usepackage{multirow}
\usepackage{xcolor}
\usepackage[]{natbib}
\usepackage{graphicx}
\graphicspath{ {./images/} }

\title{Evaluating the effects of preprocessing, method selection, and hyperparameter tuning on SAR-based flood mapping and water depth estimation}

\author{
 Jean-Paul Travert \\
  EDF R\&D, Laboratoire National d’Hydraulique et Environnement (LNHE), Chatou, France\\
  Laboratoire d’Hydraulique Saint-Venant (LHSV), ENPC, Institut Polytechnique de Paris, EDF R\&D,
Chatou, France\\
  \texttt{jean-paul.travert@edf.fr} \\
  \And
 Cédric Goeury\\
  EDF R\&D, Laboratoire National d’Hydraulique et Environnement (LNHE), Chatou, France\\
  Laboratoire d’Hydraulique Saint-Venant (LHSV), ENPC, Institut Polytechnique de Paris, EDF R\&D,
Chatou, France\\
   \And
 Sébastien Boyaval \\
  Laboratoire d’Hydraulique Saint-Venant (LHSV), ENPC, Institut Polytechnique de Paris, EDF R\&D,
Chatou, France\\
  Inria, Paris, France\\
  \And
 Vito Bacchi\\
  EDF R\&D, Laboratoire National d’Hydraulique et Environnement (LNHE), Chatou, France\\
\And
   Fabrice Zaoui\\
  EDF R\&D, Laboratoire National d’Hydraulique et Environnement (LNHE), Chatou, France\\
}

\begin{document}
\begin{justify}
\maketitle

\begin{abstract}
Flood mapping and water depth estimation from Synthetic Aperture Radar (SAR) imagery are crucial for calibrating and validating hydraulic models. This study uses SAR imagery to evaluate various preprocessing (especially speckle noise reduction), flood mapping, and water depth estimation methods. The impact of the choice of method at different steps and its hyperparameters is studied by considering an ensemble of preprocessed images, flood maps, and water depth fields.  

The evaluation is conducted for two flood events on the Garonne River (France) in 2019 and 2021, using hydrodynamic simulations and in-situ observations as reference data. Results show that the choice of speckle filter alters flood extent estimations with variations of several square kilometers. Furthermore, the selection and tuning of flood mapping methods also affect performance. While supervised methods outperformed unsupervised ones, tuned unsupervised approaches (such as local thresholding or change detection) can achieve comparable results. The compounded uncertainty from preprocessing and flood mapping steps also introduces high variability in the water depth field estimates.

This study highlights the importance of considering the entire processing pipeline, encompassing preprocessing, flood mapping, and water depth estimation methods and their associated hyperparameters. Rather than relying on a single configuration, adopting an ensemble approach and accounting for methodological uncertainty should be privileged. For flood mapping, the method choice has the most influence. For water depth estimation, the most influential processing step was the flood map input resulting from the flood mapping step and the hyperparameters of the methods. 
\end{abstract}


\section{Introduction}

Flood risk management largely benefits from accurate, timely, and spatially extensive observations of flood events. Satellite remote sensing allows the monitoring of large areas with increasing spatial and temporal resolution. Among the various types of satellite sensors, Synthetic Aperture Radar (SAR) sensors are particularly valuable for flood monitoring \citep{oberstadler1997assessment,bates2012integrating}. Unlike optical sensors, SAR sensors acquire images regardless of cloud coverage and daylight, making them suitable for detecting flood events with high spatial resolution \citep{tarpanelli2019potential}. Flood mapping with SAR imagery exploits the interaction between the emitted signal and water surfaces, which typically results in dark spots in the images due to specular reflection where the smooth water surface mirrors the signal away from the sensor \citep{hostache2009water,mason2009flood,martinis2010automatic}.

Numerous methods have been developed to extract flood maps from SAR imagery, including global and local thresholding methods \citep{chini2017hierarchical,mason2012automatic}, active contour models \citep{horritt1999statistical}, change detection methods \citep{giustarini2012change,bovolo2007split} or supervised classification approaches \citep{bentivoglio2022deep,mateo2021towards,bonafilia2020sen1floods11}. These flood maps provide essential data for model calibration and validation \citep{di2009technique,montanari2009calibration}, and data assimilation in hydraulic forecasting systems \citep{hostache2010assimilation,lai2014variational,giustarini2011assimilating,dasgupta2021mutual}. 

Beyond generating flood maps, estimating water depth from SAR data is also valuable for hydraulic model calibration and validation \citep{hostache2009water,schumann2007high,betterle2024water}. These estimations usually involve combining flood maps and ancillary datasets, such as Digital Elevation Models (DEMs) \citep{hostache2009water,schumann2007high,betterle2024water} or outputs from hydrodynamic simulations \citep{brown2016progress}. 

However, extracting hydraulic information such as flood maps and water depth fields from SAR images is subject to various sources of uncertainty. These include measurement noise (e.g., speckle), terrain-induced distortions, and vegetation and infrastructure influence. Furthermore, uncertainty arises from methodological choices in the processing workflow, such as noise filtering strategy, flood mapping methods, and hyperparameter settings. Previous studies attempted to quantify these uncertainties. For example, \cite{schumann2008estimating}
propagated geolocation uncertainties to derive ensembles of flood maps and associated water depth. Similarly, \cite{martinis2015comparing} compared various operational flood mapping strategies to study their relevance in operational contexts. \cite{landuyt2018flood} compared flood mapping methods with hyperparameter tuning across multiple flood events in Ireland and the UK (comparison to Copernicus Emergency
Mapping Service flood maps), highlighting that method performance was highly variable depending on the study case. They also underlined that the variability in flood extent maps outputs due to hyperparameters was important for some method choices. However, flood studies rarely analyze the influence of preprocessing or hyperparameter tuning across the complete workflow, including flood mapping and water depth estimation. For instance, \cite{landuyt2018flood} assume a unique preprocessing strategy for the satellite images without evaluating its role in the subsequent analysis.

In this study, we proposed a workflow for SAR-based flood analysis, evaluating the sensitivity of flood mapping and water depth estimation to different combinations of preprocessing, flood mapping, and water depth estimation methods. The uncertainty introduced at each step of the SAR image processing is quantified to identify robust configurations for operational use. The preprocessing (speckle filtering), flood mapping, and water depth estimation methods were all evaluated for varying hyperparameter settings. This evaluation was carried out in an operational context with two flood events on the Garonne River in France in 2019 and 2021, with two Sentinel-1 SAR observations available for both flood events. All code and data are publicly available to facilitate reproducibility and further experimentation by the hydraulic modeling community at \url{https://github.com/jtravert/sar-flood-evaluation-framework}.

This article is structured as follows. Section \ref{sec:material} describes the methodology and study area, including the satellite data and validation datasets. Section \ref{sec:preprocessing} describes and applies the SAR image preprocessing steps, explicitly focusing on speckle filtering strategies. Section \ref{sec:floodmapping} reviews and evaluates flood mapping approaches. Section \ref{sec:depthestimation} presents the estimation of water depths from SAR-derived flood maps using DEMs. Section \ref{sec:discussion} discusses limitations of this study and the operational use for model calibration. Finally, Section \ref{sec:conclusion} provides conclusions and implications for operational flood monitoring.

\section{Methodology}\label{sec:material}

The study presents a workflow for processing Synthetic Aperture Radar (SAR) images to support hydraulic information extraction, specifically for generating flood maps and water depth fields. The main objective was to investigate the influence on the final outputs (flood maps and water depth fields) of different method choices, including their associated hyperparameters. Although validation datasets, such as hydrodynamic simulations, were used to support a better understanding of the methodology, the main focus of the study was to explore the variability in outputs resulting from these choices, rather than to perform a strict validation. Each combination of preprocessing, flood mapping, and water depth estimation methods listed in Table \ref{table:all_methods} was evaluated. Each stage of the workflow generates a set of outputs that feed into the next stage, resulting in an ensemble of flood maps and water depth fields.
\begin{table}[ht!]
\centering
\resizebox{\textwidth}{!}{\begin{tabular}{lllll}
\hline
Category & Method/Filter & Hyperparameter & Hyperparameter values \\
\hline
\multirow{7}{*}{Speckle filtering} 
& No preprocessing & - & - \\
& Median & Window size & \{3;~5;~7\} \\
& Lee & Window size & \{3;~5;~7\} \\
& Lee Sigma & Window size & \{3;~5;~7\} \\
&  & Cumulative probability ($\xi$) & \{0.7;~0.8;~0.9\} \\
& Frost & Window size & \{3;~5;~7\} \\
&  & Damping factor ($\alpha$) & \{1;~2;~3\} \\
& Deep-learning-based (SAR2SAR) & Model weights & Pre-trained model weights (fixed) \\
\hline
\multirow{9}{*}{Flood mapping} 
& Global thresholding & Threshold selection procedure & \{Otsu, Kittler and Illingworth\} \\
& Local thresholding & Minimum tile size in pixels & \{$100\times100$;~$200\times200$\} \\
&  & Ashman's D threshold & \{1.9;~2.0;~2.1\} \\
&  & Bhattacharyya coefficient & \{0.98;~0.99\} \\
&  & Surface ratio & \{0.05;~0.1;~0.15\} \\
& Active contour & Contour smoothness ($\alpha$) & \{0.05;~0.1;~0.2;~0.3;~0.4;~0.5\} \\
& Change detection & Classification method & Global or local thresholding \\
& Supervised classification & Model & CNN or Random Forest (fixed weights) \\
& Morphological operations & Holes area in pixels & \{10;~50;~100\} \\
&  & Patches removal in pixels & \{10;~50;~100\} \\

\hline
\multirow{5}{*}{Water depth estimation} 
& Fw-DET & Slope threshold & \{no threshold;~5\%;~10\%\} \\
&  & Number of smoothing iterations & \{3;~5;~10\} \\
& FLEXTH & Slope threshold & \{no threshold;~5\%;~10\%\} \\
&  & Maximum number of neighbors & \{5;~10;~20\} \\
& Cross-section analysis & - & - \\

\hline
\end{tabular}}
\caption{Overview of speckle filtering, flood mapping, and water depth estimation methods used in this study, along with their associated hyperparameter sampling.}
\label{table:all_methods}
\end{table}

The workflow for processing SAR satellite images for flood applications is illustrated in Figure \ref{fig:workflow} and consists of the following steps: 
\begin{enumerate}
    \item \textbf{SAR image preprocessing:} Raw SAR images are first preprocessed to improve flood signal extraction. Five filtering methods were tested, including the Median filter, Lee filter \citep{lee1980digital}, Lee Sigma filter \citep{lee2008improved}, Frost filter \citep{frost1982model}, and the SAR2SAR deep-learning-based approach \citep{dalsasso2021sar2sar}. Each speckle filtering method was applied with a specific set of hyperparameters, resulting in 26 unique configurations: one configuration without any preprocessing (no hyperparameters), three configurations each for the Median and Lee filters, nine configurations each for the Frost and Lee Sigma filters, and one configuration for SAR2SAR. 
    
    Section \ref{sec:preprocessing} provides a detailed overview of the preprocessing step, and Table \ref{table:all_methods} presents the different configurations.
    \item \textbf{Flood mapping:} Each preprocessed image is used as input for flood mapping. Five methods were tested: global thresholding, local thresholding, active contour model, change detection, and supervised classification. These methods are also parameterized, resulting in an ensemble of flood maps for each preprocessed input. Post-processing morphological operations are optionally applied to remove isolated water pixels or small non-physical holes caused by speckle, geometric distortions, or flood map processing. In total, for each input preprocessed image, 48 flood mapping configurations are tested: one configuration each for the supervised classification models (no hyperparameters), two for global thresholding, two for change detection, six for active contour, and 36 for local thresholding. When using morphological post-processing, nine configurations were evaluated. With morphological operations, 432 flood mapping outputs (48 $\times$9) were generated per preprocessed image.
    
    A more detailed explanation of the flood mapping step can be found in Section \ref{sec:floodmapping}, along with the flood mapping configurations in Table \ref{table:all_methods}.
 
    \item \textbf{Water depth estimation:} Each flood map generated in the previous step, together with a Digital Elevation Model (DEM), is used to estimate water depth fields using three methods: Fw-DET \citep{cohen2019floodwater}, FLEXTH \citep{betterle2024water}, and a cross-sectional hydraulic approach. For each input flood map, 19 water depth estimation configurations are tested: one configuration for the cross-section approach (no hyperparameters), nine for Fw-DET, and nine for FLEXTH. 

    A comprehensive discussion of this step is presented in Section \ref{sec:depthestimation}, and the water depth estimation configurations are presented in Table \ref{table:all_methods}. 

\end{enumerate}
 The influence of method selection and hyperparameter choices is studied for the outputs from preprocessing, flood mapping, and water depth estimation. The range of the hyperparameters is based on classical values used in the literature to avoid non-physical hyperparameter values.  

\begin{figure*}[ht!]
    \centering
    \includegraphics[width=0.95\linewidth]{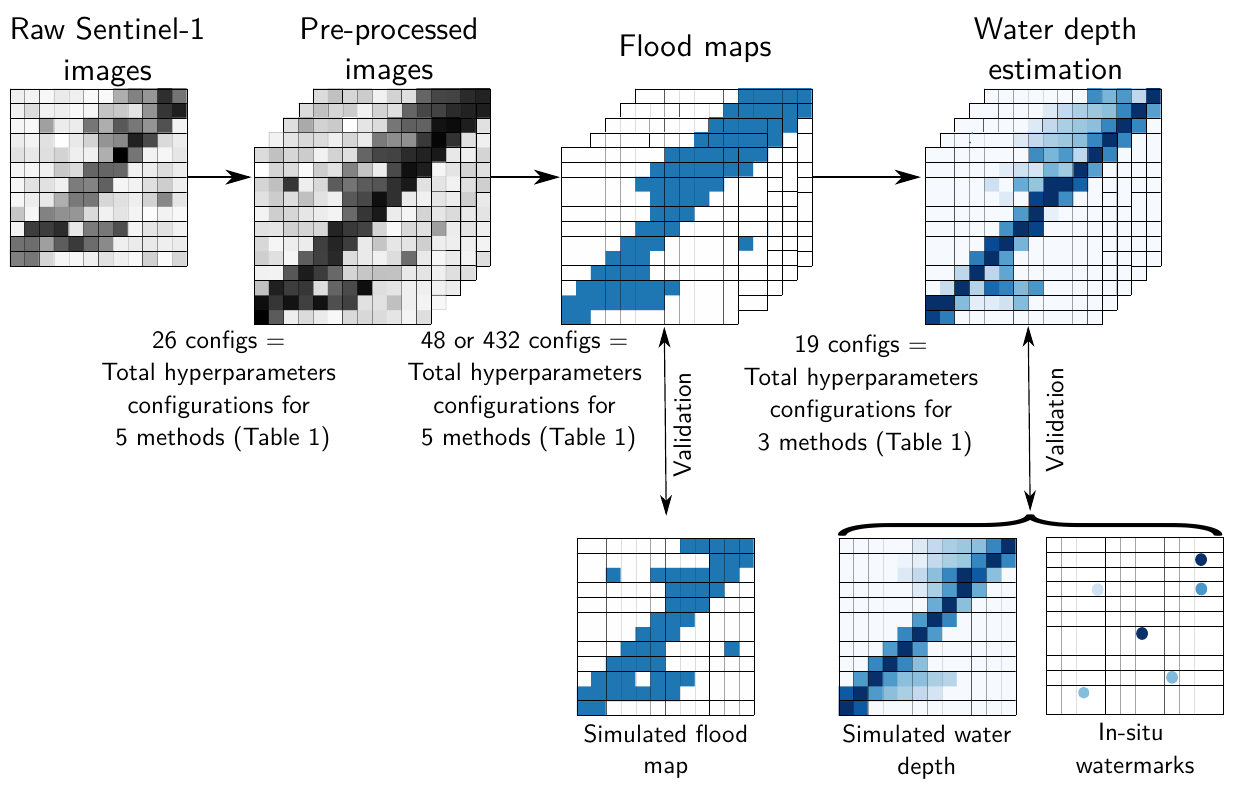}
    \caption{Workflow for processing Sentinel-1 (SAR) images to derive flood maps and water depth fields. Raw images are preprocessed using 26 hyperparameter configurations, then each image is turned into 48 flood maps (or 432 if using morphological operations). Each flood map is transformed into water depth fields using 19 configurations. The configurations are described in Table \ref{table:all_methods}.}
    \label{fig:workflow}
\end{figure*}


\subsection{Study area and materials}

The study area was the Garonne River between Tonneins and La Réole, in southwest of France (see Figure \ref{fig:studyareasatobs}). In this section of the river, the river width is around 250 m and the floodplain is 1-4 km wide, mainly comprises rural areas used for agricultural purposes. Since the end of the nineteenth century, the area has been equipped with dikes to protect urban areas and infrastructures. The floodplain, aside from the presence of dikes, displays minimal topographic variation.

\subsubsection{Satellite observations}
The study area was observed during two flood events in December 2019 and early February 2021  by the Sentinel-1 C-band Synthetic Aperture Radar (SAR) instrument at
5.405 GHz. The extent of the study area and the satellite image acquisitions are visualized in Figure \ref{fig:studyareasatobs}. Sentinel-1 ``Ground Range Detected'' products were downloaded from ASF Data Search Vertex (\url{https://search.asf.alaska.edu/}) in both $VH$ and $VV$ polarizations. The role of polarization is beyond the scope of this study. Most flood mapping methods rely on the analysis of a single polarization image. Accordingly, the $VH$ polarization was used since, in the literature, $VH$ polarization better distinguishes flooded from dry areas than $VV$ polarization \citep{henry2006envisat}. 
Two additional Sentinel-1 images, acquired under non-flooded conditions, were used as references for one of the flood mapping methods. For the 2019 event, the reference image was acquired on 10 December 2019. For the 2021 event, the reference observation was acquired on 28 January 2021. Each image consists of an $N_{x}\times N_{y}$ grid of pixels, with a spatial resolution of $10 \times 10$ m. The images are projected onto a common grid that covers the entire study area. Here, the image's dimensions are $N_{x}=2644$ and $N_{y}=2312$. All times reported below are given in Coordinated Universal Time (UTC). 

\begin{figure*}[ht!]
    \centering
    \includegraphics[width=0.7\linewidth]{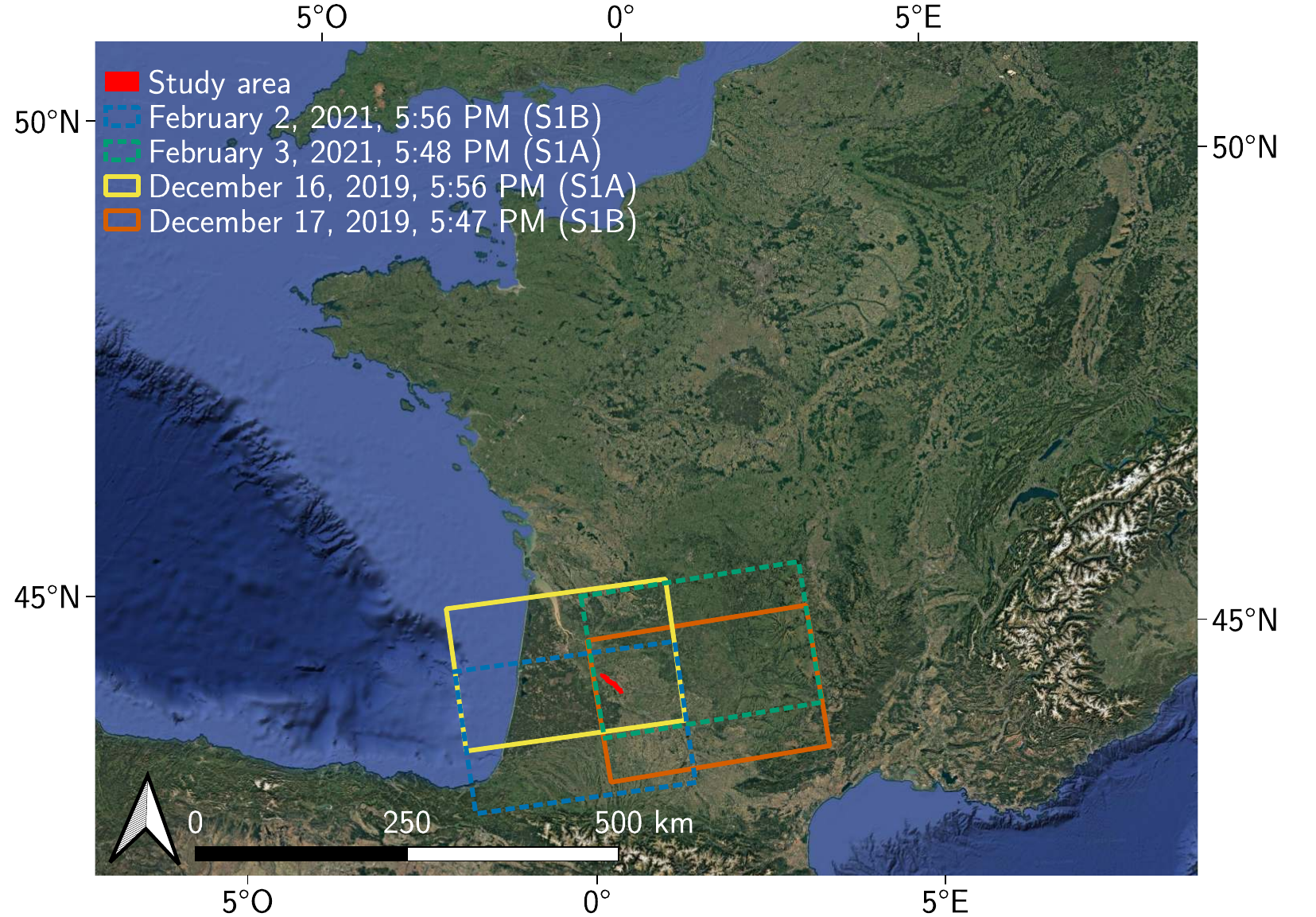}
    \caption{Visualization of the study area on the Garonne River in France and the extent of the Sentinel-1 A (S1A) and Sentinel-1 B (S1B) acquisitions during 2019 and 2021 flood events.}
    \label{fig:studyareasatobs}
\end{figure*}

\subsubsection{Validation data}\label{sec:validationdata}
For comparing the outputs of the observed flood maps and water depth fields, observed water marks and stage gauging stations are available. We also compare the outputs to simulated flood maps and water depth fields. The simulations are not the ground truth, but serve as a reference for comparison. The main goal of the study is to compare the variability of the outputs due to preprocessing, method choices, and hyperparameters, so the validation dataset is not the most important. We describe these datasets below.

Watermarks are visible traces left on buildings, trees, or other infrastructures during a flood event at the peak water level. For both the 2019 and 2021 flood events, watermarks were collected and made available on the French collaborative platform ``Repères de Crues'' (\url{https://www.reperesdecrues.developpement-durable.gouv.fr/}). For the 2019 and 2021 flood events, 121 and 178 are available, respectively. For both events, satellite images were acquired near the flood peak (17 December 2019 and 3 February 2021), so the watermarks should coincide with the water depths extracted from the satellite images. 

Three stage-gauging stations are available in the study area (Tonneins, Marmande, and La Réole). Discharge and water level data at these locations during the flood events are available on Vigicrues  (\url{https://www.vigicrues.gouv.fr/}, a flood-monitoring service that collects watermarks during floods in France). The measured discharges at the three stations for both flood events are shown in Figure \ref{fig:dischargeWD}. 
\begin{figure*}[ht!]
    \centering
    \includegraphics[width=1\linewidth]{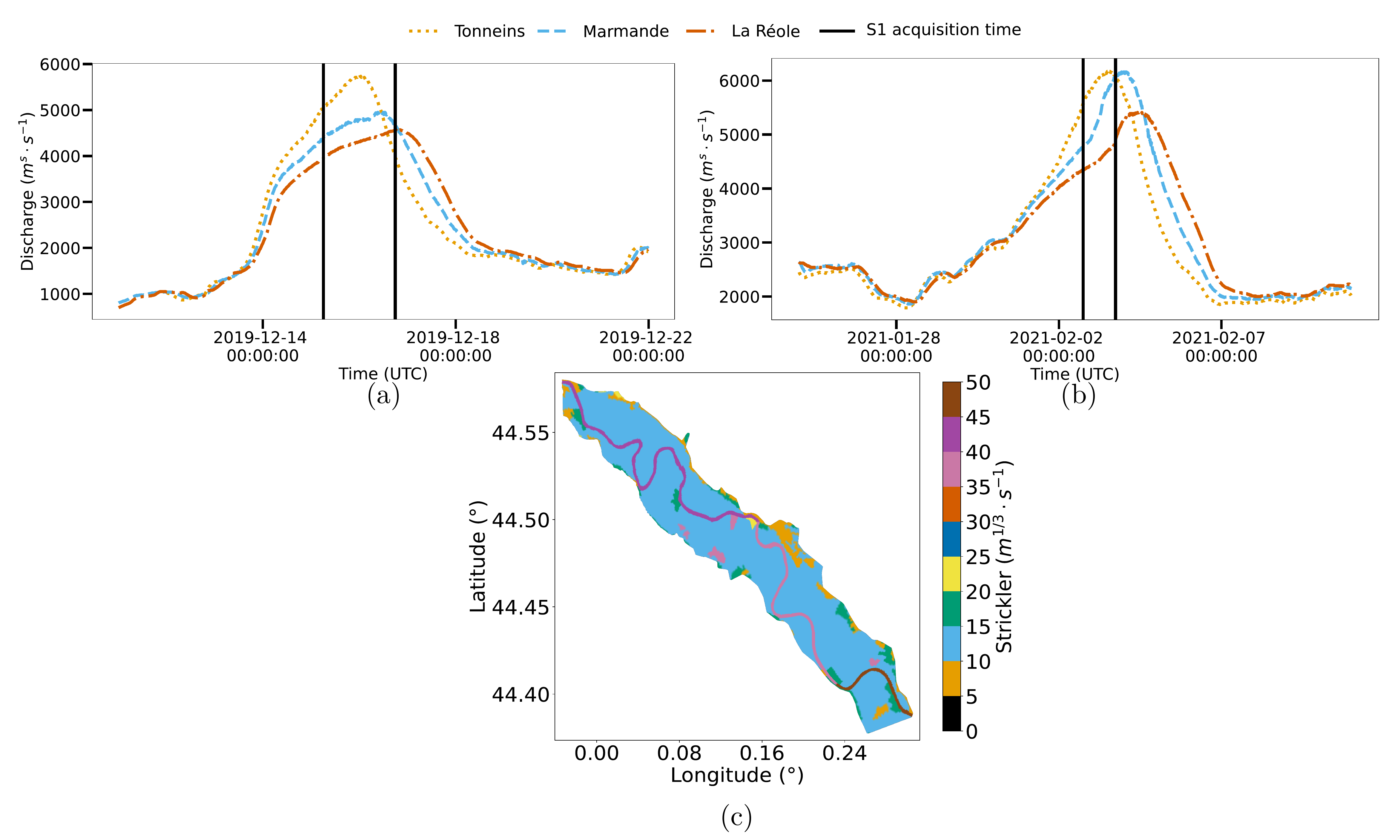}
    \caption{Measured discharge at gauging stations (Tonneins, Marmande, and La Réole) during the (a) December 2019 and (b) February 2021 flood event, along with (c) spatially distributed Strickler values used in the study area.}
    \label{fig:dischargeWD}
\end{figure*}\\

For flood simulations, a numerical solution of the Shallow Water Equations (SWEs) was used and solved with TELEMAC-2D, part of the openTELEMAC open-source hydrodynamic modeling system (www.opentelemac.org) \citep{hervouet2007hydrodynamics}. The SWEs are expressed in Cartesian coordinates where gravity acts uniformly in the vertical direction as $-g\overrightarrow{e_z}$: 

\begin{align}\label{eq:swe}
  \frac{\partial h}{\partial t}+\frac{\partial (hu)}{\partial x}+\frac{\partial (hv)}{\partial y}&=0,\\
     \frac{\partial (hu)}{\partial t}+\frac{\partial(hu^{2})}{\partial x}+\frac{\partial (huv)}{\partial y}&=-gh\frac{\partial \eta}{\partial x}+\nabla \cdot (h \, \nu_{e}\nabla u)-\frac{1}{\rho}\tau_{bx},\\ 
          \frac{\partial (hv)}{\partial t}+\frac{\partial(hv^{2})}{\partial y}+\frac{\partial (huv)}{\partial x}&=-gh\frac{\partial \eta}{\partial y}+\nabla \cdot (h \, \nu_{e}\nabla v)-\frac{1}{\rho}\tau_{by}.  \label{eq:thirdSWE}
\end{align}
The system unknowns are the water depth $h\ge 0$ and the depth-averaged velocity $u\overrightarrow{e_x}+v\overrightarrow{e_y}$ both functions are defined as function of spatial coordinates $x$, $y$ and time $t\in [0,T)$.
The water surface is denoted by $\eta=h+z_b$, where $z_b(x,y)$ is the prescribed bottom elevation. A constant viscosity $\nu_{e}>0$ is assumed, and the bed shear stress is expressed as
$\tau_{bx}\overrightarrow{e_x}+\tau_{by}\overrightarrow{e_y}$ depending on the variables $h$, $u$, and $v$. The bed shear stress is computed using Manning-Strickler formulation \citep{manning1890flow}:
\begin{equation}
   \left\{
    \begin{array}{ll}
\tau_{bx}=\frac{\rho\cdot g\cdot u}{h^{1/3}\cdot K_{s}^{2}}\sqrt{u^{2}+v^{2}} \\[2ex]
\tau_{by}=\frac{\rho \cdot g\cdot v}{h^{1/3}\cdot K_{s}^{2}}\sqrt{u^{2}+v^{2}}
    \end{array}
\right.
,\label{eq:fricT2D}
\end{equation}
where $K_{s}$ is the Strickler coefficient which varies spatially with $x$, and $y$ \citep{morvan2008concept}.

Upstream discharge was retrieved from the Tonneins stations for 11-21 December 2019 and 25 January-10 February 2021 and reported in Figure \ref{fig:dischargeWD}. A rating curve was imposed at the downstream boundary. The simulations were initialized at $t=0$ with a base flow of 800 $m^{3}/s$ for the 2019 event, and 2300 $m^{3}/s$ for the 2021 event. One simulation was conducted for each flood event. The Strickler values were calibrated in previous studies for the river channel \citep{besnard2011comparaison,el2019uncertainty,nguyen2022dual}, and are fixed in the floodplains based on the land use and tabulated values in the literature \citep{chow1988applied} as described in Table \ref{table:roughness}. The spatial distribution of Strickler values used in the simulation is reported in Figure \ref{fig:dischargeWD}.

\begin{table}[ht!]
\caption{Strickler friction values used in the floodplains.}
\label{table:roughness}
\centering
 \begin{tabular}{l c}
 \hline
      Land use & Strickler value ($m^{1/3}\cdot s ^{-1}$)\\
 \hline
      Waterbodies & 35 \\
      
      Fields and meadows without crops & 20 \\
      Cultivated fields with low vegetation &17.5\\
      Cultivated fields with high vegetation & 12.5 \\
      Shrublands and undergrowth areas& 10\\
      Areas of low urbanization & 9 \\
      Highly urbanized areas & 6.5\\
      \hline
    \end{tabular}
\end{table}

As this article focuses on processing satellite observations, the numerical model is not detailed. The reader can refer to \cite{besnard2011comparaison} and \cite{travert2024} for more information on the construction and parameterization of the numerical model.

\section{Satellite images preprocessing}\label{sec:preprocessing}
Synthetic Aperture Radar (SAR) images, such as those acquired by Sentinel-1, require several preprocessing steps to correct for sensor geometry, normalize radiometric responses, and suppress speckle noise. The processing chain adopted in the present study is presented in Figure \ref{fig:chap4:preprocessing} and comprises sequential operations: application of orbit files, removal of thermal and border
noise, radiometric calibration, speckle filtering, and terrain correction using the
Range-Doppler method. In the processing workflow for terrain correction and georeferencing, a Digital Elevation Model of the study area at a 1 m resolution was used. Two alternative workflows were used depending on the speckle filtering strategy since the deep-learning-based speckle filtering (SAR2SAR method) was trained on images without preprocessing, while the other traditional filters were applied on calibrated images. 
\begin{figure*}[ht!]
    \centering
    \includegraphics[width=1\linewidth]{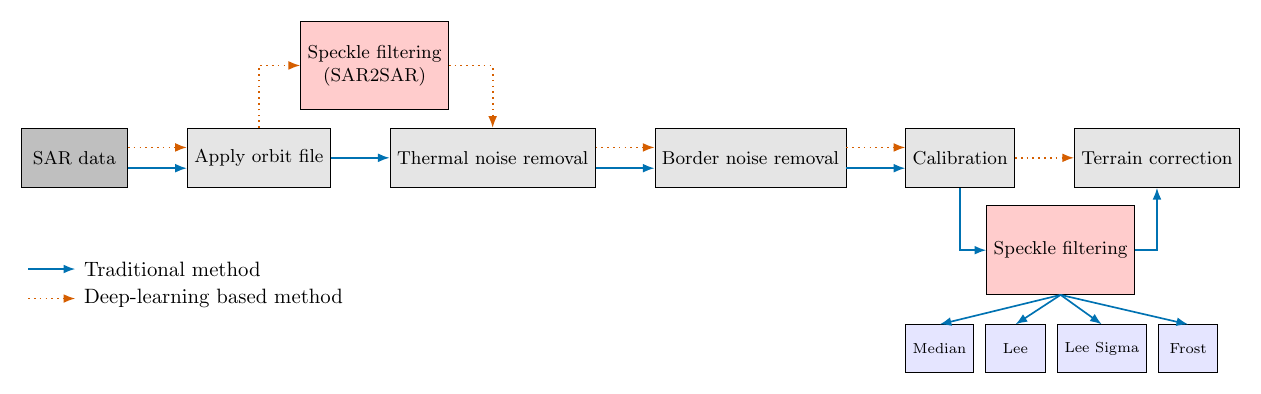}
    \caption{Preprocessing workflow for Sentinel-1 SAR images.}
    \label{fig:chap4:preprocessing}
\end{figure*}

The main source of error in SAR imagery is speckle noise, which arises from the coherent summation of scattered electromagnetic waves. It resembles salt and pepper with dark and bright pixels (see Figure \ref{fig:specklenoise}). It alters the statistical properties of the image, preventing them from maintaining a consistent mean radiometric level in homogeneous areas \citep{bruniquel1997multi}. Speckle noise is modeled with a randomly fluctuating variable \citep{goodman1976some}, such as:
\begin{equation}\label{eq:multiplicativenoise}
I_{i,j} = R_{i,j} \cdot S_{i,j},
\end{equation}
where \( I \in \mathbb{R}^{N_{x}\times N_{y}}\) denotes the observed intensity (raw image), \(  R \in \mathbb{R}^{N_{x}\times N_{y}} \) the true radar backscatter, \(  S \in \mathbb{R}^{N_{x}\times N_{y}} \) the speckle component, and $i,j$ the pixel locations. $S_{i,j}$ and $R_{i,j}$ are assumed to be statistically independent. 

\subsection{Speckle filters methods}
To mitigate the effects of noise, several speckle filters that aim to recover  \( R \) the true radar backscatter while preserving image structures have been proposed in the literature. The choice of filters and their hyperparameters affects the interpretability of the image and the accuracy of downstream tasks such as flood mapping. A review on speckle filtering methods is available in \cite{deledalle2014nl} and \cite{lee2017polarimetric}. In this study, we selected a representative set of widely used traditional filters, such as Frost, Lee, Lee Sigma, and Median filters, due to their simplicity and strong track record in SAR image denoising. Additionally, we included a recent deep learning-based method to evaluate potential performance improvements from modern data-driven approaches. The characteristics of these filters are summarized in Table \ref{table:all_methods}. In the following sections, each method is described in more detail.

The deep-learning based model (SAR2SAR model) was implemented via the deepdespeckling Python library available at \url{https://pypi.org/project/deepdespeckling/} \citep{dalsasso2021sar2sar}. The other traditional filtering operations were conducted using the ESA SNAP toolbox.  The complete preprocessing chain was automated using SNAP’s Graph Processing Tool (GPT) and Python scripting, facilitating large-scale processing of SAR image stacks. 


\subsubsection{Median filter}
The Median filter replaces the center pixel with the median value of all pixels within a local scanning window, such that:
\begin{equation}
  \hat{R}_{i,j} = \operatorname{median} \left\{I_{m,n} \;\middle|\; i - k \leq m \leq i + k,\; j - k \leq n \leq j + k \right\},
\end{equation}
with a window size of $2k+1$ with $k$ a non-negative integer. In this study, the window size is treated as a hyperparameter. The Median filter is effective at removing isolated spot noise, but it tends to blur edges and erase thin linear features \citep{lee1983digital}. To address this limitation, other approaches referred to as adaptive filters have been developed that incorporate local image statistics to better preserve structural details.

\subsubsection{Lee filter}
The Lee filter \citep{lee1980digital} accounts for the local statistics of the image within a moving window. From the multiplicative noise definition in Equation \eqref{eq:multiplicativenoise}, the mean and variance of $R$ can be estimated, such as:

\begin{equation}
\bar{R}_{i,j} = \bar{I}_{i,j}, \quad \quad
\mathbb{V}ar(R_{i,j}) = \frac{\mathbb{V}ar(I_{i,j}) - \bar{I}_{i,j}^2 \mathbb{V}ar({S_{i,j}})}{\mathbb{V}ar({S_{i,j}})+ \bar{S}_{i,j}^{2}},
\end{equation}
where the quantities $\bar{I}_{i,j}$ and $\mathbb{V}ar(I_{i,j})$ are the local mean and variance of pixel intensities within a window of size $2k+1$ centered at $(i,j)$. In this study, the window size is treated as a hyperparameter. The Lee filter assumes a linear estimator of the form $\hat{R}=a\bar{R}+bI$, where $\hat{R}$ is the minimum mean square estimate of $R$, and $a$ and $b$ are constants to minimize the mean square error. Then, using the local mean and variance within each scanning window, the estimator is written as:
\begin{equation}
    \hat{R}_{i,j}=\bar{R}_{i,j}+\frac{\mathbb{V}ar(R_{i,j})}{\mathbb{V}ar(I_{i,j})}(I_{i,j}- \bar{S}_{i,j}\bar{R}_{i,j}).
\end{equation}
The Lee filter aims to reduce speckle in homogeneous areas while preserving image details, such as edges and fine structures, in areas of high variance.

\subsubsection{Lee Sigma filter}
The Lee filter effectively reduces speckle noise but can cause blurry edges and loss of detail in heterogeneous areas. To address this issue, the Lee Sigma filter \citep{lee1983digital} was introduced. It calculates local statistics similarly to the Lee filter but using only the pixels whose intensities fall within a range to exclude outliers. The  range is defined as  $[\bar{R}_{i,j}-2\sqrt{\mathbb{V}ar({S_{i,j}})}\bar{R}_{i,j},\bar{R}_{i,j}+2\sqrt{\mathbb{V}ar({S_{i,j}})}\bar{R}_{i,j}]$. Because the a priori mean, $\bar{R}_{i,j}$, is unknown, it is approximated by $I_{i,j}$, the value of the center pixel.

An improved version \citep{lee2008improved} relaxes the usual range and introduces a new interval $(I_1, I_2)$ that meets two conditions:
\begin{itemize}
    \item The interval captures a fixed cumulative probability $\xi$:
    \begin{equation}
        \xi = \int_{I_1}^{I_2} p(I)\, dI,
    \end{equation}
    where $p(I)$ is the empirical probability distribution of the pixel intensity computed on the scanning window. 

    \item The mean within the interval must match the overall mean:
    \begin{equation}
        \bar{I} = \frac{1}{\xi} \int_{I_1}^{I_2} I\, p(I)\, dI.
    \end{equation}
\end{itemize}
A value of $\xi=0.8$ or 0.9 is often used, but a lower value of $\xi$ may be selected to preserve SAR image texture information and prevent potentially over-smoothing details \citep{lee2008improved}. Two hyperparameters are used for the Lee Sigma filter, the window size and $\xi$. 

\subsubsection{Frost filter}
Similarly to the Lee filter, the Frost filter \citep{frost1982model} is based on the local statistics of the images and the multiplicative noise model. The Frost filter replaces a pixel value with a weighted sum of the values of its neighbors within a moving scanning window, such that:
\begin{equation}
    \hat{R}_{i,j} = \sum_{m=-k}^{k} \sum_{n=-k}^{k} w_{m,n} \cdot I_{i+m, j+n},
\end{equation}
where the weights $w_{m,n}$ decrease with distance from the pixel of interest with an exponential decay controlled by a damping factor $\alpha$, such that:
\begin{equation}
w_{m,n} = \frac{e^{-\alpha \, d(m,n)}}{\sum_{m=-k}^{k} \sum_{n=-k}^{k} e^{-\alpha \, d(m,n)}},
\end{equation}
with $d(m,n)$ the Euclidean distance of pixels at position $m,n$ from the pixel at the center of the scanning window. For a pixel center located at $i,j$ the distance is defined as $d(m,n)=\sqrt{(m-i)^{2}+(n-j)^{2}}$. For the Frost filter, two hyperparameters are considered in this study, the window size and the damping factor $\alpha$.
\subsubsection{Deep-learning based filtering}
Due to the difficulty of removing speckle noise with empirical models based on image statistics, deep learning approaches are increasingly used. SAR2SAR \citep{dalsasso2021sar2sar} is a deep learning-based despeckling method using a U-Net architecture \citep{ronneberger2015u} which aims to predict speckle noise. The U-Net is trained with synthetic speckle realizations, then fine-tuned on real SAR image pairs from a time series. After training, weights are used to compute the denoised image as a function of input pixel intensities. In this study, we used a pre-trained model with weights available at \url{https://gitlab.telecom-paris.fr/ring/sar2sar}. For SAR2SAR, no hyperparameters were tested.

\subsection{Application of speckle filtering}
We applied the five filtering methods for all hyperparameter configurations to the four Sentinel-1 SAR images (and to the reference non-flooded images). In our implementation, the computation time for the speckle filtering step ranged from a few seconds for the Lee and Median filters, to approximately 20-30 seconds for Lee Sigma and Frost filters, and up to around 4 min for SAR2SAR method, on Intel(R) Core(TM) i7-11850H @ 2.50GHz processor. Visual inspection and quantitative evaluations were carried out to compare the outputs of each method and their variability due to hyperparameter settings. This section analyzes the results for the $VH$ polarization, but the results for $VV$ polarization were similar. Figure \ref{fig:specklenoise} illustrates how these filters (for one hyperparameter configuration) influence the backscatter for the same Sentinel-1 image.

\begin{figure*}[ht!]
    \centering
    \includegraphics[width=0.9\linewidth]{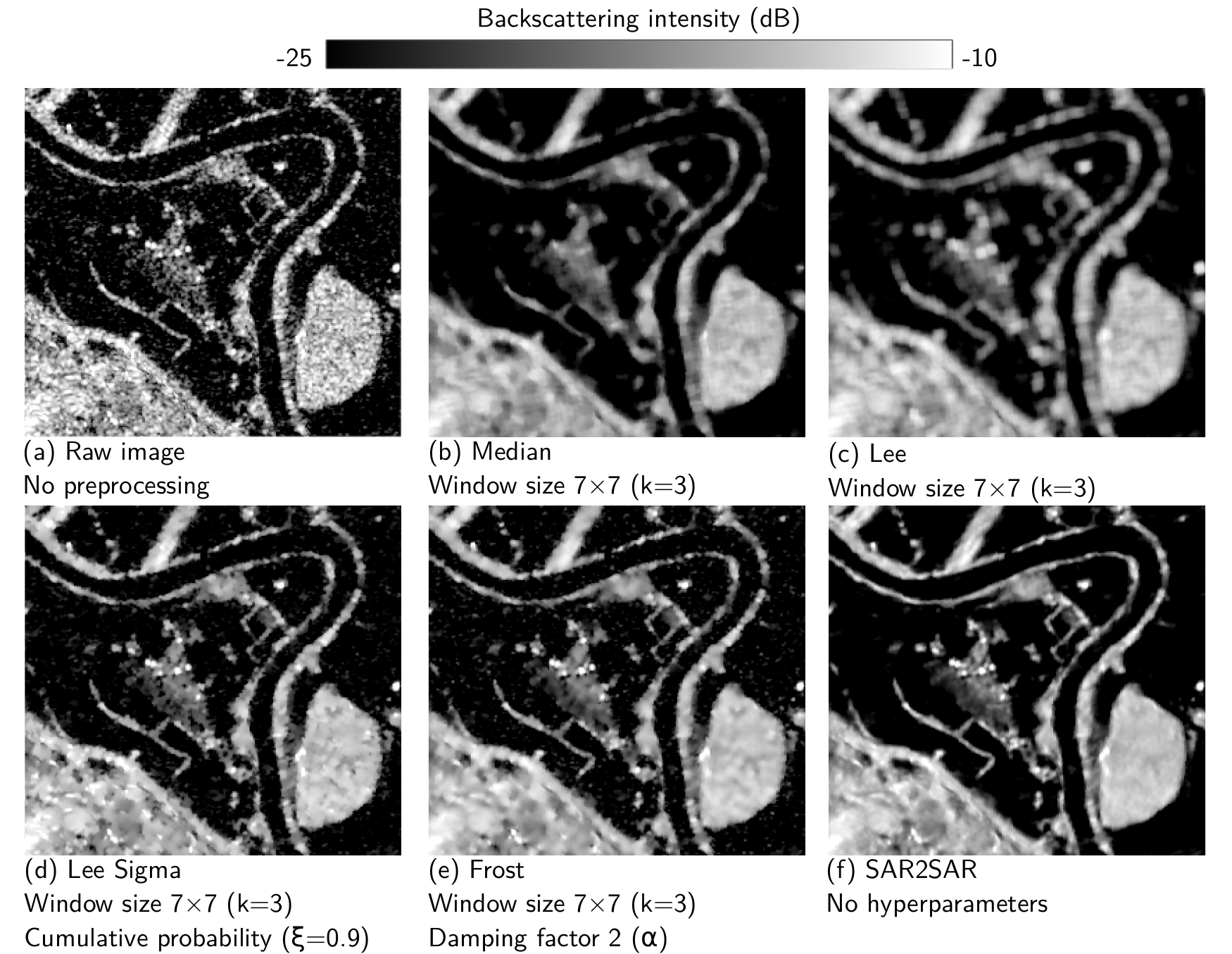}
    \caption{Backscatter intensity results from applying various speckle filters on a Sentinel-1 image in VH polarization.}
    \label{fig:specklenoise}
\end{figure*}

The raw Sentinel-1 image (see Figure \ref{fig:specklenoise}a) exhibited significant speckle noise, appearing as salt and pepper granular regions. All tested filter configurations reduced the noise to varying extents, with variable success for preserving the edges. The Median and Lee filters (see Figure \ref{fig:specklenoise}b-c) reduced speckle noise but significantly blurred structural details along curvilinear structures. The Lee Sigma and Frost filters (see Figure \ref{fig:specklenoise}d-e) better balance noise reduction and detail preservation, effectively preserving both edges and textures. The SAR2SAR approach (see Figure \ref{fig:specklenoise}f) reduced the speckle noise while visually preserving edge features. Visually, for these hyperparameter configurations and on this zone, the SAR2SAR approach seemed to outperform the traditional methods.

The amount of speckle reduction can be quantified by calculating the Equivalent Number of Looks (ENL) over a quasi-homogeneous area, defined as:

\begin{equation}
    \mathrm{ENL} = \frac{\bar{\hat{R}}^2}{\mathbb{V}ar({\hat{R}})},
\end{equation}
where $\bar{\hat{R}}$ is the mean estimated intensity over a homogeneous area, and $\mathbb{V}ar({\hat{R}})$ is the variance of the estimated intensity over that same area. ENL estimates the signal-to-noise ratio, and the higher the ENL, the better the speckle suppression. 

\begin{figure*}[ht!]
    \centering
    \includegraphics[width=0.9\linewidth]{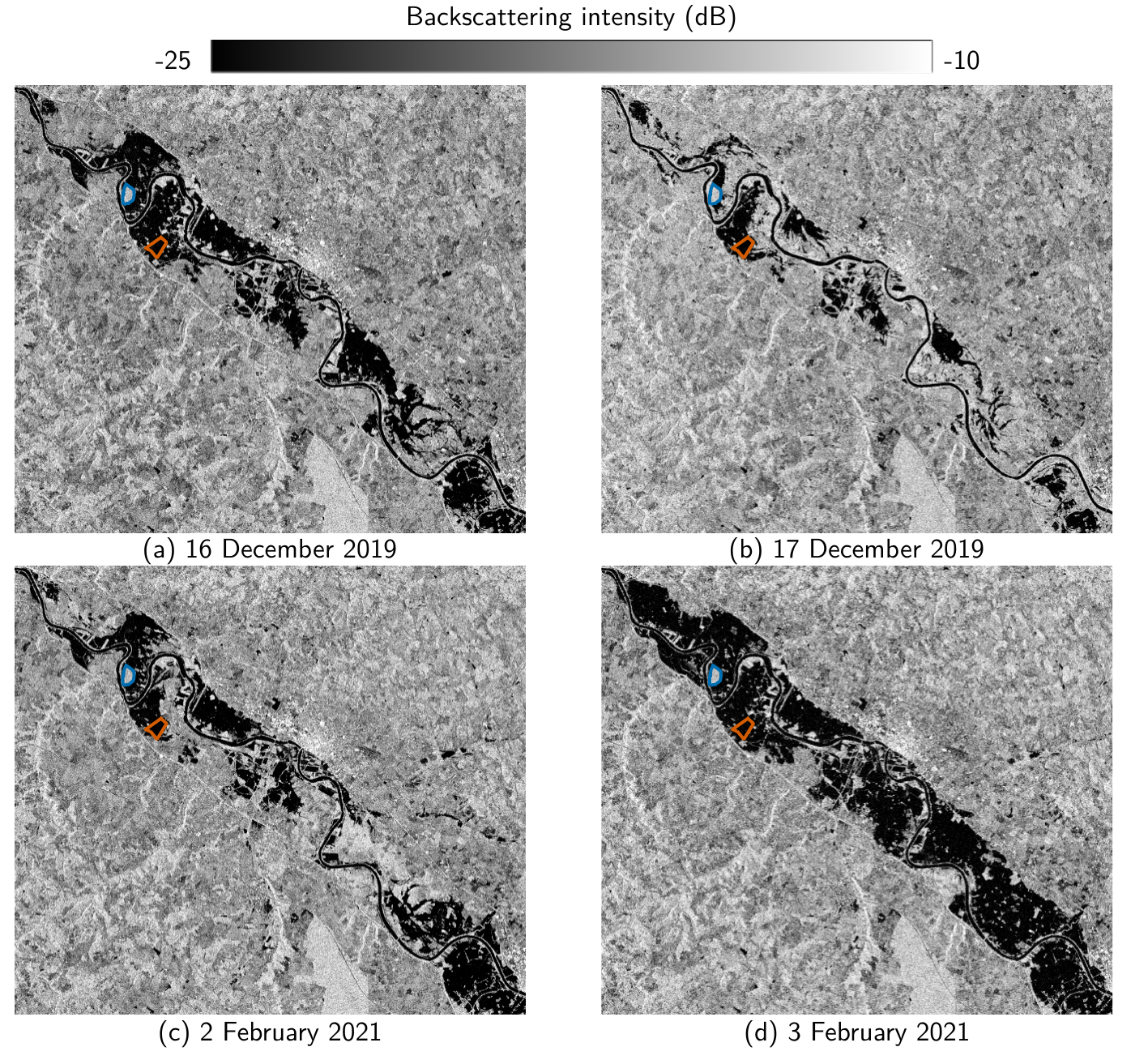}
    \caption{Four raw Sentinel-1 images of the 2019 and 2021 flood events on the Garonne River in France. The blue and orange polygons are used for the Equivalent Number of Looks computation.}
    \label{fig:ENLzones}
\end{figure*}

The ENL was computed for two zones, including a dry vegetated region and flooded region in blue and orange, respectively, in Figure \ref{fig:ENLzones}, for all the preprocessed images (corresponding to 26 configurations in total according to Table \ref{table:all_methods} plus one configuration without filtering) and for the four satellite images. The results are reported in Figure \ref{fig:ENLresults}, where each point corresponds to the ENL of the zone for one of the preprocessed images. All filtering methods contributed to speckle reduction, as indicated by the systematically higher ENL values than the images without preprocessing. The SAR2SAR achieved the highest ENL values across all dates and regions, particularly for the dry region (see Figure \ref{fig:ENLresults}a). Traditional statistical filters, such as the Lee and Frost filters, also improved ENL for some hyperparameter configurations (largest window size and higher damping factor). The high variability of the ENL (e.g., from 10 to 30 for Figure \ref{fig:ENLresults}a) for the Lee, Median, and Frost filters underlined the significant impact of the method's hyperparameters on speckle reduction. 

\begin{figure*}[ht!]
    \centering
    \includegraphics[width=1\linewidth]{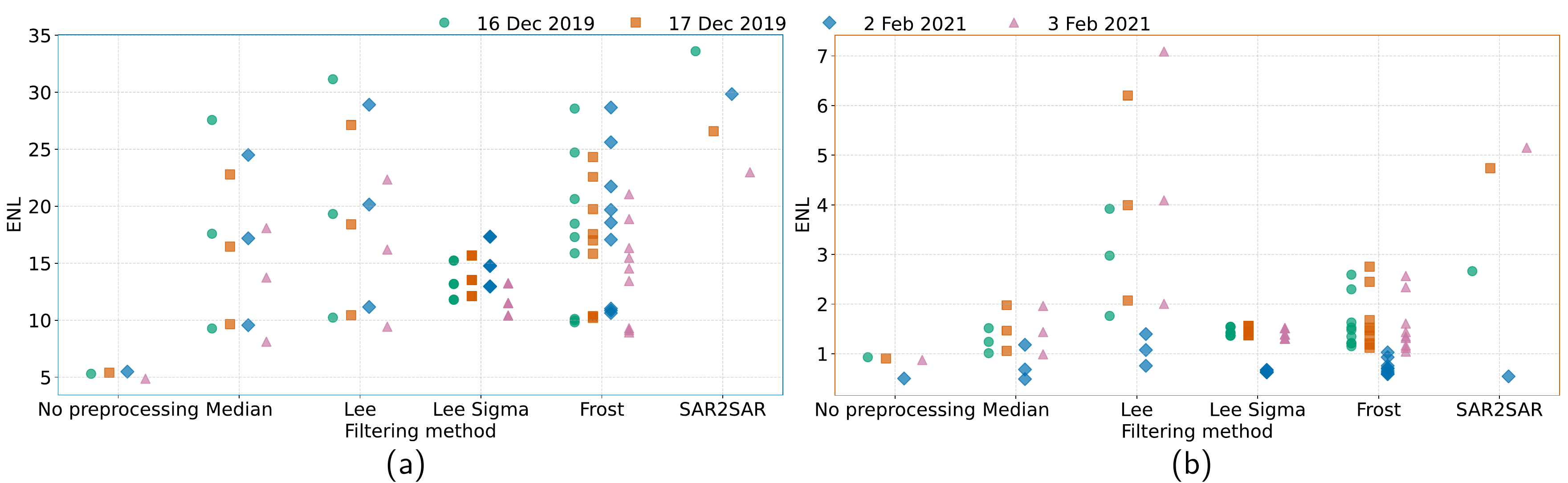}
    \caption{Equivalent Number of Looks (ENL) for the four satellite images with different preprocessing for the dry homogeneous area (a) and flooded homogeneous area (b).}
    \label{fig:ENLresults}
\end{figure*}

The quantitative analysis of edge and structure preservation is difficult because fine structures are close to the speckle noise spatial resolution \citep{lee1994speckle}. Edge preservation was not quantitatively evaluated in this study. The SAR2SAR approach gave the best quantitative results on noise reduction in homogeneous areas while visually preserving edges and structures. Traditional filtering methods had similar ENL values for some hyperparameter settings, showing their potential to reduce speckle noise but at the cost of preserving details (edge blurring) and the need for hyperparameter tuning. 

Speckle reduction aims to improve the interpretability of the images and the extraction of information. Thus, the role of speckle reduction is analyzed in more detail in Section \ref{sec:floodmapping} and \ref{sec:depthestimation} to show its impact on interpreting SAR images and how the variability in speckle reduction changes the output. 

 \FloatBarrier

\section{Flood mapping}\label{sec:floodmapping}
In this study, we generated an ensemble of flood maps by applying several flood mapping methods across a range of hyperparameters, listed in Table \ref{table:all_methods}. The methodology for flood map generation is illustrated in Figure \ref{fig:methodologyfloodmapping}, and follows this general workflow:

\begin{enumerate}
    \item \textbf{Input:} Preprocessed SAR images $\hat{R}$. Only $VH$ polarization images are used except for supervised classification, which uses both $VH$ and $VV$ polarizations.
    \item \textbf{Method selection:} Global thresholding, local thresholding, active contour model, change detection, or supervised classification (Convolution Neural Networks or Random Forest). Each method is evaluated over a range of hyperparameters (see Table \ref{table:all_methods}). For instance, two methods for evaluating the threshold parameter are tested for global thresholding.
    \item \textbf{Flood map generation:} Each parameter configuration results in a flood map $O\in\{0,1\}^{N_{x}\times N_{y}}$
    \item \textbf{Post-processing (optional):} Morphological operations are applied on the flood maps to fill small holes and remove small elements not connected to the flood.
\end{enumerate}

\begin{figure*}
    \centering
    \includegraphics[width=1\linewidth]{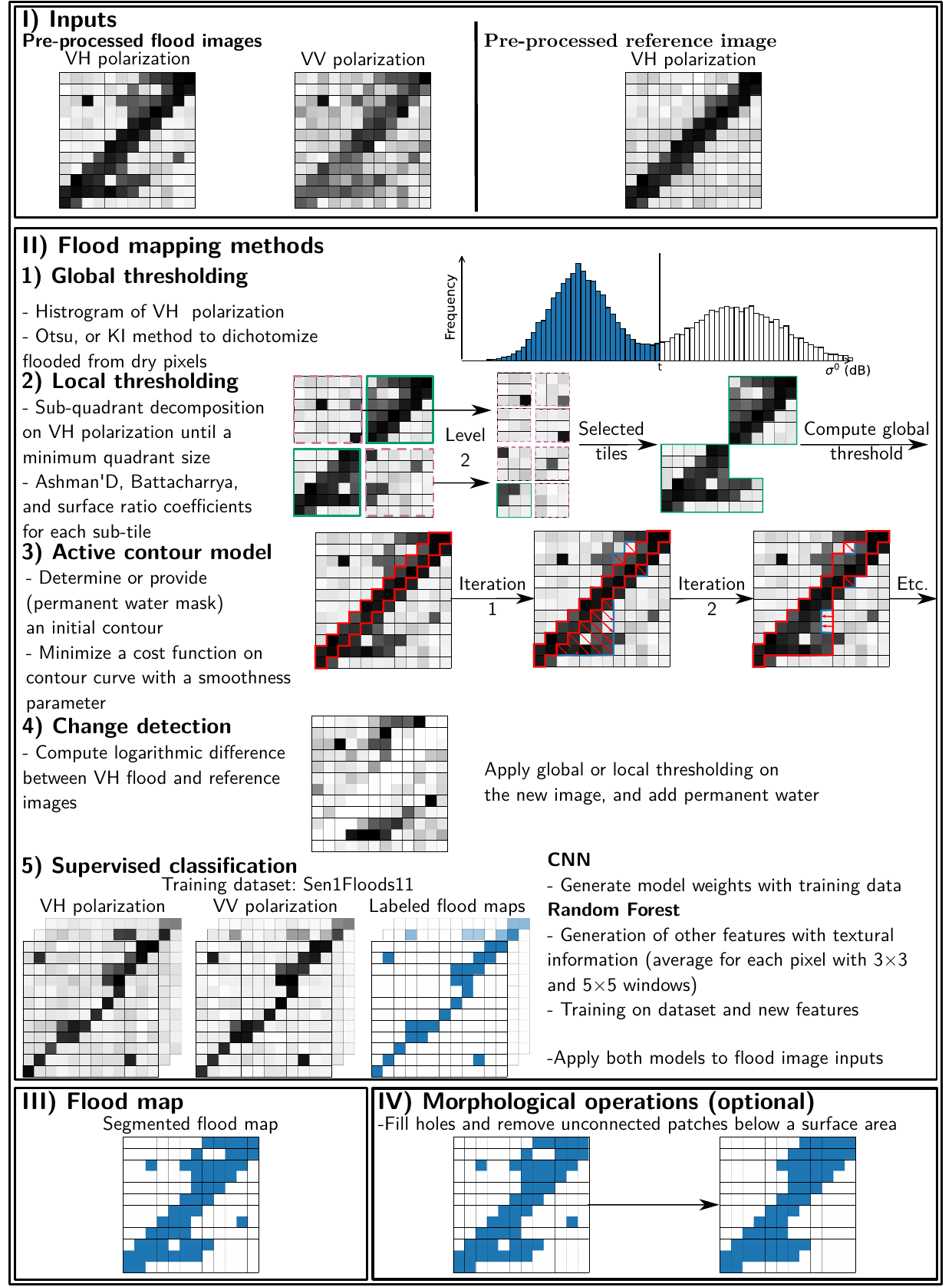}
    \caption{Schematic overview of the flood mapping methodology.}
    \label{fig:methodologyfloodmapping}
\end{figure*}

\subsection{Methods}

\subsubsection{Global thresholding}\label{sec:globalthresholding}

Global thresholding of SAR images is a straightforward method to separate foreground (e.g., flooded pixels) from background (e.g., dry pixels). This approach assumes that the two-pixel classes (e.g., flooded and dry pixels) can be separated with a threshold value for the whole image. The threshold value is usually determined automatically with various methods \citep{sezgin2004survey}.

In this study, we evaluated the impact on the flood mapping of two widely used global thresholding methods: Otsu's method \citep{otsu1979threshold} and the Kittler and Illingworth (KI) method \citep{kittler1986minimum}, each briefly described below.
\begin{itemize}
    \item Otsu's method selects the threshold that maximizes the between-class variance, which separates the histogram of pixel intensities into two distinct classes. The between-class variance is defined as:
\begin{equation}
     \sigma_{B}^{2}=\omega_{f} \omega_{b}(\mu_{f}-\mu_{b})^{2},
 \end{equation}
 where $\mu_{f}$ and $\mu_{b}$ are the mean intensities of the background and foreground classes, respectively. $\omega_{f}$ and $\omega_{b}$ are the foreground and background class fractions, respectively. 
 \item The Kittler and Illingworth (KI) method also considers a mixture of two Gaussian distributions. The optimal threshold separates both classes by minimizing a cost function that quantifies the overlap between the distributions and is calculated as follows:
\begin{equation}
    J = 1 + 2 \left( \omega_{f} \log(\sigma_{f}) + \omega_{b} \log(\sigma_{b}) \right) - 2 \left( \omega_{f} \log(\omega_{f}) + \omega_{b} \log(\omega_{b}) \right),
\end{equation}
where $\sigma_{f}$ and $\sigma_{b}$ are the standard deviations of the foreground and background classes, respectively.
\end{itemize}


Global thresholding methods are widely used in flood mapping for their simplicity \citep{martinis2009towards,schumann2010near,pulvirenti2013discrimination}. However, they assume that the histogram is bimodal, which may not hold when the number of flooded pixels is low or due to spatial variability in backscatter caused by terrain or surface conditions across the image.

\subsubsection{Local thresholding}\label{sec:localthresholding}
To address the limitations of global thresholding, local-based thresholding was introduced \citep{martinis2015fully,twele2016sentinel,chini2017hierarchical}.  Images are divided into smaller tiles to identify bimodal histograms representing flooded and dry classes. From these identified sub-tiles, a threshold is computed with global thresholding. Following \cite{chini2017hierarchical}, we used quad-tree decomposition to recursively split the image until reaching a minimum tile size (to guarantee statistical representativeness). A tile is eligible for thresholding if its histogram is bimodal, the distributions are normally distributed, and both classes are sufficiently represented. These conditions are evaluated by computing three qualitative coefficients:
\begin{itemize}
    \item Ashman's D coefficient \citep{ashman1994detecting}, which quantifies the separation between two Gaussian distributions, and is defined as:
\begin{equation}
        AD = \frac{|\mu_f - \mu_b|}{\sqrt{0.5(\sigma_{f}^2 + \sigma_{b}^2)}}.
\end{equation}
    \item Bhattacharyya Coefficient \citep{bhattacharyya1943measure}, which quantifies the amount of overlap between two distributions, and is computed as:
    \begin{equation}
    BC=\sum_{i}\sqrt{hist(y_{i})}\sqrt{hist_{f}(y_{i})},
    \end{equation}
    where $y$ is the pixel values in the tiles, $hist$ is the histogram of the distribution, $hist_{f}$ is the fitted histogram with two Gaussian curves, and $i$ is the bins of the two discrete histograms.
    \item The surface ratio is defined as the ratio of the area (measured in number of pixels) covered by the smaller population (e.g., dry or flooded pixels) to that of the larger population.
\end{itemize}

This study evaluated the impact on flood maps of the minimum tile size and the thresholds on the Ashman D, Bhattacharyya coefficient, and surface ratio for different values reported in Table \ref{table:all_methods}. For instance in \cite{chini2017hierarchical}, $AD$ should be above 2, $BC$ above 0.99 and the surface ratio above $10\%$.

\subsubsection{Active contour models}
A classical active contour model is based on the Chan-Vese segmentation method \citep{chan2001active}, which is inspired by the Mumford–Shah model \citep{mumford1989optimal}. This approach determines a contour $C$ that segments the image into two regions so that the pixel intensities are approximately constant within each region. The segmentation problem is formulated as the following minimization:
\begin{equation}
\min_{C} \ 
\alpha \, \text{Length}(C) + \nu \, \text{Area}(\text{inside}(C)) 
+ \lambda_1 \sum_{(i,j) \in \text{inside}(C)} (\hat{R}_{i,j} - c_1)^2 
+ \lambda_2 \sum_{(i,j) \in \text{outside}(C)} (\hat{R}_{i,j} - c_2)^2,
\end{equation}
where $\alpha \geq 0$ controls the smoothness of the contour, 
 $\nu \geq 0$ penalizes the area inside the contour,
 $\lambda_1, \lambda_2 > 0$ are weighting the data fidelity inside and outside the contour, respectively, and $c_1$ and $c_2$ are the average intensities inside and outside the contour $C$, respectively. The algorithm is adapted so that $c_{1}<c_{2}$, forcing flooded pixels with lower backscattering to be inside $C$.
 
 In this study, we considered $\alpha$ as a hyperparameter, while the remaining parameters were fixed as follows for computational cost reasons: $\lambda_1 = 2$, $\lambda_2 = 1$, and $\nu = 0$. We chose to constrain the problem with $\lambda_{1}>\lambda_{2}$ in order to favor the identification of regions with lower mean intensities (typically associated with flooded areas in SAR imagery) as the foreground while assigning higher-intensity regions to the background. Active contour models require an initial contour as a starting condition for the segmentation process. \cite{horritt1999statistical} suggests selecting the initial contour manually. The simplest method is to initialize the contour using a mask of permanent water bodies. Maps of permanent water bodies are available on the Global Flood Monitoring Service with a 10 m resolution (\url{https://global-flood.emergency.copernicus.eu/}) \citep{martinis2022towards}. 

\subsubsection{Change detection}
Change detection methods require at least two satellite images of the same area, one during the flood and the others before or after the event. The output of change detection methods depends on the reference image without floods, so the reference image should be chosen carefully \citep{hostache2012change}. Although there are methods using a stack of images \citep{clement2018multi}, we focused on a pair of images. First, we computed the logarithmic difference between the two images because of the multiplicative character of speckle noise (see Equation \eqref{eq:multiplicativenoise}) \citep{bazi2005unsupervised}. Flooded pixels are then classified using global thresholding described in Section \ref{sec:globalthresholding} with consistent parameter sampling. Similarly to global thresholding, the hyperparameter in the change detection method is the method to find the threshold on the image (Otsu or KI). 

\subsubsection{Supervised classification}
With the growing number of satellite observations, new datasets have become available. For instance, the Sen1Floods11  dataset \citep{bonafilia2020sen1floods11} includes raw Sentinel-1 imagery labeled for flood applications. These datasets are often leveraged to train machine learning methods for supervised classification of floods \citep{bentivoglio2022deep}. Random Forests (RF) and Convolutional Neural Networks (CNN) are widely used methods for classification tasks based on labeled training data \citep{zhao2020design,bentivoglio2022deep}, and are described below: \begin{itemize}
    \item Random Forest classifier \citep{breiman2001random} consists of an ensemble of decision trees, each trained on a subset of the data. The state of the pixels is classified by aggregating the predictions of all trees with majority voting. For Sentinel-1 data, the pixel intensity on both VV and VH polarizations is used as a feature. With only two features, the prediction capability of Random Forest is limited. We added textural information by providing the mean value of each pixel with window sizes of 3, 5, and 7 pixels for both polarizations, resulting in 8 features. The Random Forest classifier was trained for various numbers of estimators, maximum depth, and minimum leaf number with Bayesian optimization using the Optuna Python library \citep{akiba2019optuna} (\url{https://pypi.org/project/optuna/}).
    \item  Convolutional Neural Networks (CNN) \citep{lecun2015deep} exploit spatial context by applying convolutional filters over local neighborhoods of pixels to extract information. Unlike Random Forest, CNN is adapted for identifying texture in images. This spatial awareness makes CNNs particularly effective in capturing contextual information that single-pixel classifiers, such as Random Forest, may miss.
\end{itemize}

Both models are trained with the Sen1Floods11 dataset \citep{bonafilia2020sen1floods11} using $VV$ and $VH$ polarization bands and hand-labeled data dichotomizing flooded from dry regions. The trained model weights were used to generate flood maps from satellite imagery based on pixels' backscattering intensity. For reproducibility, the weights of both models are available at: \url{https://github.com/jtravert/sar-flood-evaluation-framework/tree/main/sources/1_FloodExtent/methods/MLweights}. The training procedure for the CNN was adapted from \cite{bonafilia2020sen1floods11}. For supervised classification methods, no hyperparameters are considered. 
\subsection{Post-processing of flood maps}

After applying flood mapping methods, the flood maps can contain errors caused by measurement noise, speckle, or processing errors. To improve the spatial coherence of flood maps, morphological operations (standard methods used in image processing) can be used to refine the flood maps. These operations include hole filling, which fills small gaps within flooded regions, and removal of isolated patches not physically connected to the flooded areas. 

In this study, the influence of these operations was studied for holes and small patches of water from 1,000 $m^{2}$ to 10,000 $m^{2}$. We applied morphological operations by filling holes of less than 10, 50, or 100 pixels and removing unconnected patches of less than 10, 50, or 100 pixels, representing nine configurations. The number of filled/removed pixels is the hyperparameter for morphological operations.

\subsection{Results}
Flood maps were generated using the five flood mapping methods with the hyperparameter settings detailed in Table \ref{table:all_methods}. For each satellite acquisition, 1,222 generated flood maps were produced (26 speckle filtering configurations $\times$ 48 flood mapping setups). These flood maps were optionally processed with morphological operations (nine configurations), resulting in 10,998 additional possible flood map variations for each satellite acquisition. In our implementation, the computation time for mapping a flood on a single preprocessed image was a few seconds for all methods, except for active contour, which required 5 to 10 minutes (depending on the smoothing hyperparameter), on an Intel(R) Core(TM) i7-11850H @ 2.50GHz processor. Each generated flood map was compared against hydrodynamic simulations, as described in Section \ref{sec:validationdata}. Using a pixel-to-pixel comparison approach, the evaluation focused on floodplain areas, excluding permanent water bodies. At the grid points, a hit occurs when a pixel is correctly predicted to be flooded (True Positive ($TP$)), and a correct rejection (True Negative ($TN$)) occurs when it is correctly predicted to be dry. A false alarm (False Positive ($FP$)) and a miss (False Negative ($FN$)) occur when a pixel is simulated as flooded or dry, respectively, whereas the opposite is observed. Two standard metrics for comparing flood maps \citep{hunter2005development,grimaldi2016remote} were used:

\begin{itemize}
 \item \textbf{Accuracy:} Proportion of correctly predicted pixels (both flooded and non-flooded) relative to the total number of pixels: 
\begin{equation}
    ACC= \frac{TP + TN}{TP + TN + FP + FN}.
\end{equation}  
    \item \textbf{F1-score:} The F1-score is defined as:
\begin{equation}
        F1 = \frac{2 \cdot TP}{2 \cdot TP + FP + FN}.
\end{equation}  
\end{itemize}
Both metrics range from 0 to 1, with higher values indicating better performance. 


\subsubsection{Flood mapping methods evaluation}

First, the images generated without applying morphological post-processing operations are analyzed. Figure \ref{fig:generalvariability} shows the variability of the Accuracy and F1-score metrics across the four satellite acquisitions. Each box plot represents the range of metric variations of a flood mapping method due to varying input preprocessing and hyperparameter settings. For each method, box plots for Accuracy and F1-score are displayed in different colors for the four satellite images. These box plots should be compared with one another for the same acquisition date to assess method performance variability under consistent conditions. 
\begin{figure*}[ht!]
    \centering
    \includegraphics[width=0.95\linewidth]{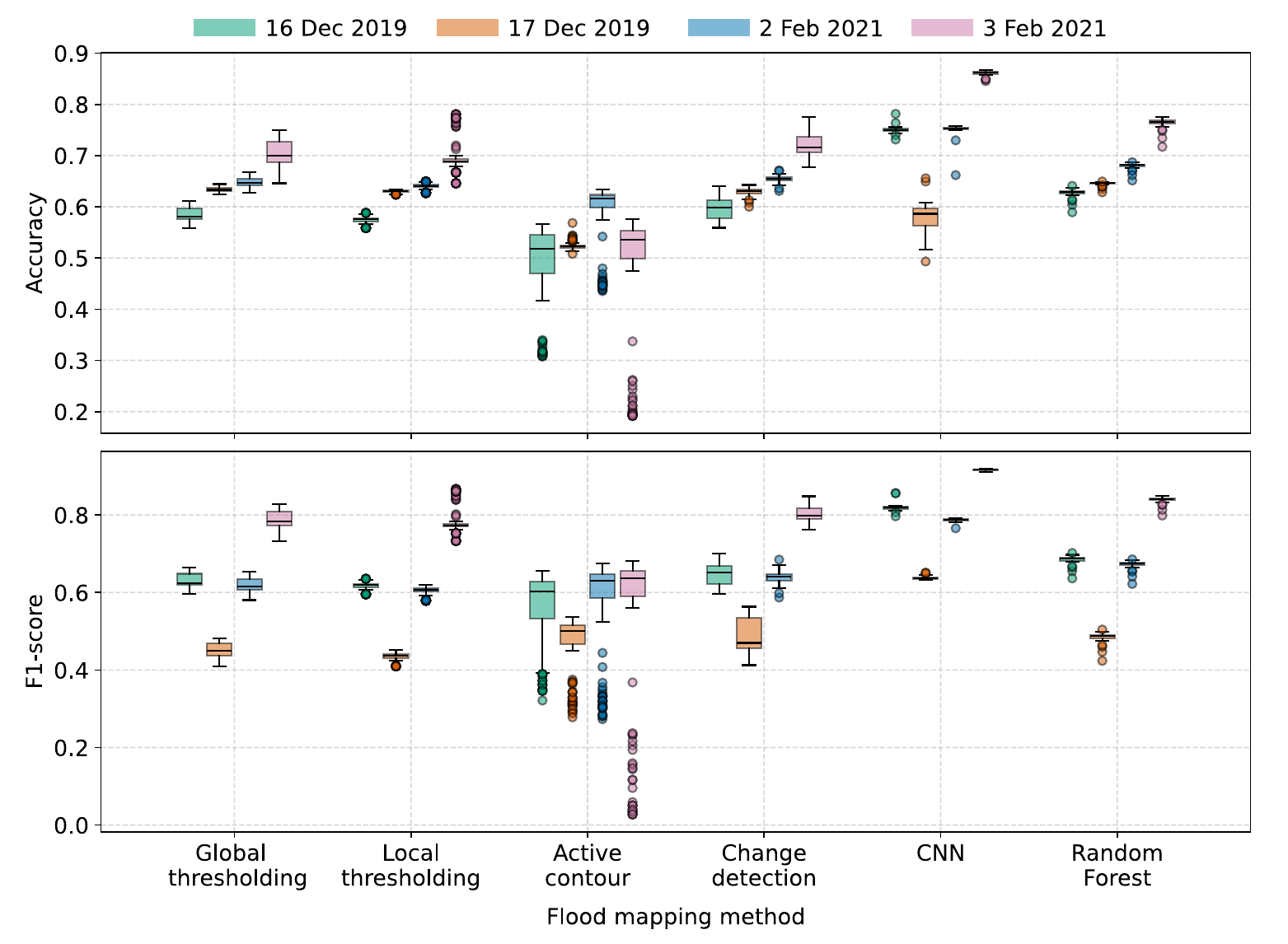}
    \caption{Accuracy and F1-score for the generated flood maps and various flood mapping methods compared against their respective simulated flood maps for four satellite images.}
    \label{fig:generalvariability}
\end{figure*}

CNN and Random Forest methods (supervised classification) outperformed unsupervised methods on median F1-score and Accuracy. The CNN approach presented the highest median performance (between 0.79 and 0.91 for F1-score) with a narrow inter-quartile range (<0.005) in three out of the four acquisition dates. The Random Forest also performed well (F1-score median between 0.67 and 0.84), but with slightly lower scores and higher variability. Both supervised classification methods exhibit low variance because of the absence of hyperparameter tuning, leaving speckle filtering as the main source of variation. In contrast, active contour exhibited larger variability in Accuracy and F1-score across all dates. It underlined the sensitivity to either the smoothing parameter ($\alpha$) or image preprocessing. While they occasionally achieved reasonable performance for specific configurations, their inconsistency (e.g., F1 range between 0.4 and 0.65 on the 16 December 2019 image) makes them less reliable in operational settings. This method can work only after careful manual tuning, limiting its operational appeal. Active contour models are usually better for local scales and simple flood patterns around the initial contour. Thresholding methods (global and local) showed moderate performance and low variability. Additional configurations with a larger hyperparameter space showed that the thresholding methods were not very sensitive to their hyperparameters. For some configurations, local thresholding resulted in comparable metric values to supervised classification while being easy to use. Finally, change detection methods provided similar moderate results regarding F1-scores and Accuracy. Their variance was relatively low but was higher for some acquisition dates (17 December 2019 and slightly 3 February 2021), likely due to differences in magnitude change between the pre- and during-flood images, with a smaller flood extent on 17 December 2019. The supervised classification method provided the best trade-off between accuracy and robustness. When training data or model weights are unavailable, local thresholding or change detection methods remain attractive because they require almost no user input and result in similar scores with appropriate tuning.

 Figure \ref{fig:floodedarea} complements the analysis of the ACC and F1-score metrics by analyzing the flooded area for all flood mapping methods and satellite acquisitions. Out of a total domain area of 135.82 km², the simulated flooded areas were 112.38 km² (3 February 2021), 83.11 km² (2 February 2021), 103.63 km² (16 December 2021), and 68.02 km² (17 December 2021). CNN produced the largest flooded area estimates, closely matching the simulated extents, though it occasionally overestimates (17 December 2019). In contrast, the other methods underestimated the flooded areas systematically relative to the simulations. In general, outputs from hydrodynamic models provide a smooth flooded surface water extent with hydraulic connectivity. However, for the generated flood maps from SAR imagery, floods in vegetated, urban areas or between the dikes are often not well detected, which can cause an underestimation of the flooded area. On the one hand, the simulations are conservative and overestimate the flooded area, while the extracted flood maps underestimate the flooded areas. However, the main goal should be to match the geometric pattern and have a similar flood map geometry between the simulations and the observations, even if there are small holes in the observed flooded area. The estimated flooded area can differ by 20 to 30 km² between methods, highlighting significant differences in flood map extraction approaches. Additionally, for a given method, the impact of hyperparameters or input images can be significant, with variations of 5 to 10 km² for active contour or change detection and CNN for 17 December 2019. 

\begin{figure*}[ht!]
    \centering
    \includegraphics[width=0.9\linewidth]{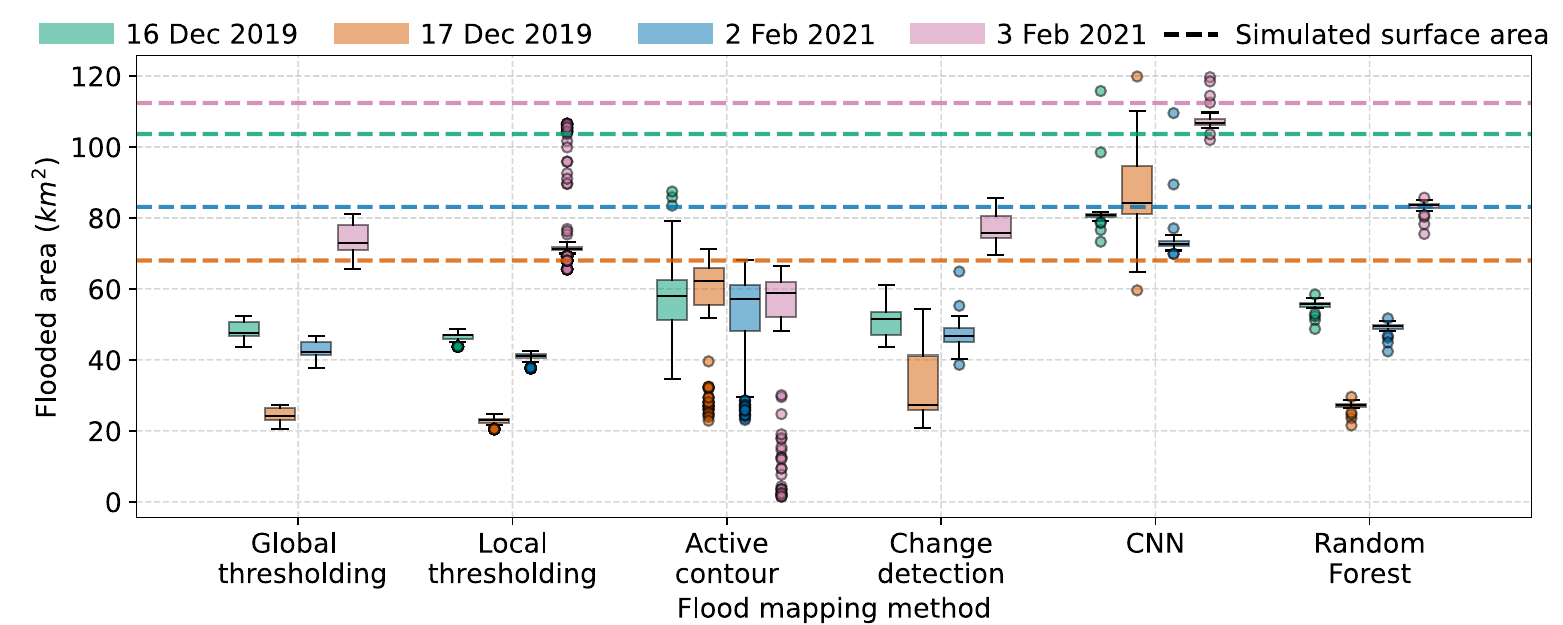}
    \caption{Flooded surface area for the generated flood maps for four satellite images and every flood mapping method.}
    \label{fig:floodedarea}
\end{figure*}

\subsubsection{Impact of preprocessing}
In most flood studies, the preprocessing of speckle noise is often assumed to be deterministic, with only one method (usually the Lee or Lee Sigma filter) \citep{di2009technique,landuyt2018flood}; however, the influence of speckle filtering choice on flood mapping can be significant. The preprocessing strategies should be evaluated for fixed configurations (flood mapping method and hyperparameter) to study the impact of preprocessing alone. For instance, the effect of preprocessing for CNN-generated flood maps is reported in Figure \ref{fig:preprocessingimpact} for the flooded surface areas. It highlighted the impact of preprocessing with a highly variable flooded area for Median filtering, depending on the window size of the filter (the only parameter for the Median filter). With filtering, the flooded area is larger than that of images without preprocessing. Between the different preprocessing methods, the flooded area has important variations showing the impact of the preprocessing method on the generated flood maps.  

\begin{figure*}[ht!]
    \centering
    \includegraphics[width=0.8\linewidth]{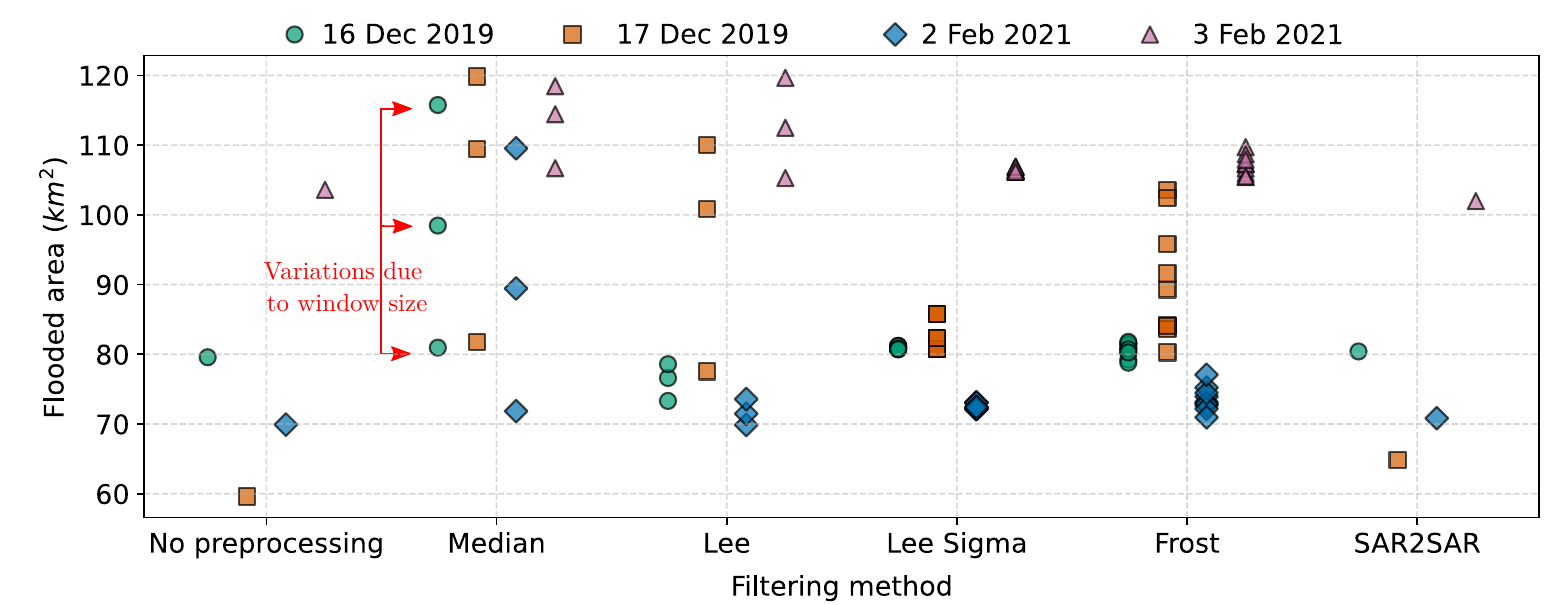}
    \caption{Impact of different filtering methods on the flooded area for flood maps generated using Convolutional Neural Networks.}
    \label{fig:preprocessingimpact}
\end{figure*}

Figure \ref{fig:preprocessingimpact} illustrates the impact of the different filtering methods on the flooded area, with a single flood mapping configuration. In total, Table \ref{table:all_methods} lists 48 configurations (2 for global thresholding and change detection, 36 for local thresholding, 6 for active contour, and one for each supervised classification method). For the Median, Lee, Lee Sigma, and Frost filters, we analyzed the impacts of their hyperparameters on these 48 configurations. The difference between the minimum and maximum value for each metric was calculated for every filter configuration (e.g., for the CNN configuration on 2 February 2021, the range of variation between the min and max values for the Median filter was 38 km²). Figure \ref{fig:preprocessingsensitivity} presents the min-max variation in F1-score and flooded area due to speckle filtering hyperparameters for 2 February 2021. Each flood mapping configuration is visualized with a point, and each has distinct markers for the different methods. The results indicate that the sensitivity to speckle filter hyperparameters varies by flood mapping method. For instance, the active contour method exhibits high sensitivity to the Median filter’s window size, with variations ranging from 7 to 23 km². In contrast, global and local thresholding methods exhibit minimal sensitivity, with variations of less than 1 km². Overall, the Lee Sigma filter exhibited the lowest variability across all configurations. Local thresholding, change detection, and Random Forest configurations are more sensitive to Lee filter or Frost filter hyperparameters than Lee Sigma or Median filters. 

\begin{figure*}[ht!]
    \centering
    \includegraphics[width=1\linewidth]{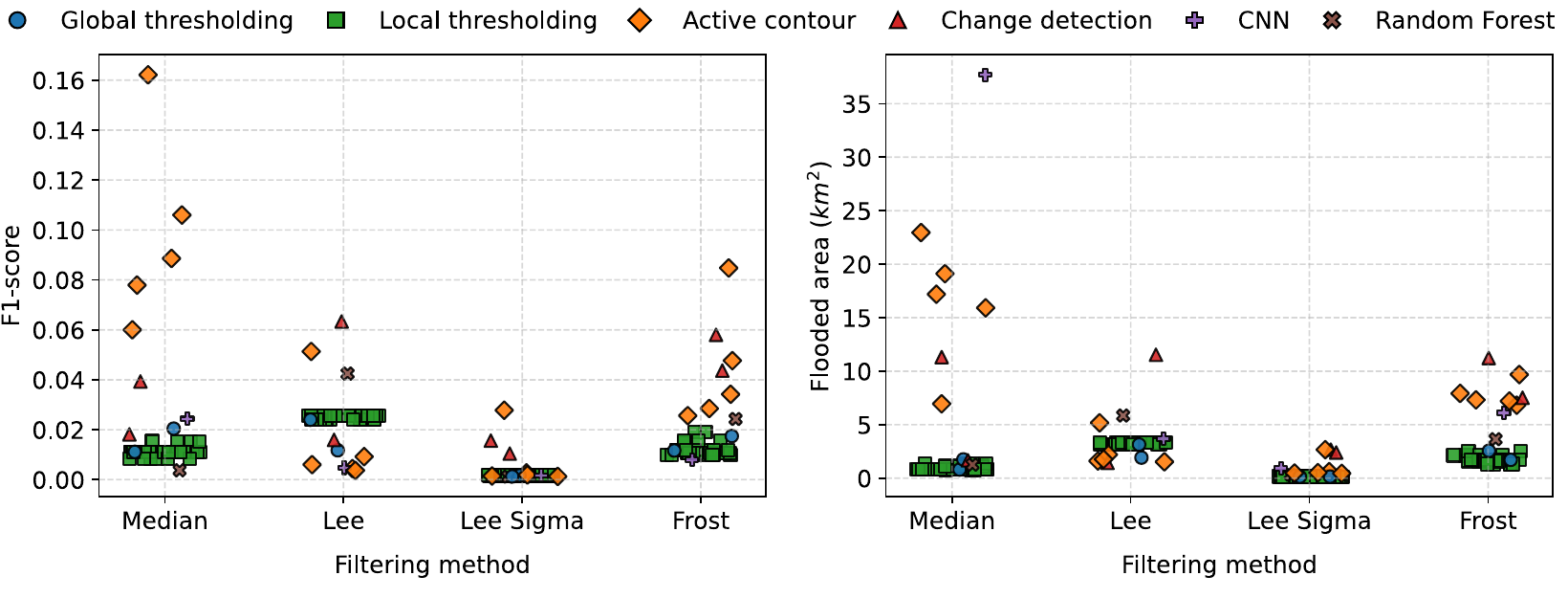}
    \caption{Range of variation (between the minimum and maximum values) of F1-score and flooded area due to filtering methods hyperparameters for the 48 flood mapping configurations for 2 February 2021 acquisition. The points are jittered horizontally for visualization purposes.}
    \label{fig:preprocessingsensitivity}
\end{figure*}

\subsubsection{Impact of hyperparameters}\label{sec:hyperparametersfloodmapping}
The role of hyperparameters used in flood mapping methods was evaluated, excluding the supervised classification methods (CNN and Random Forest), for which no hyperparameters were defined in this study. Similarly to the preprocessing analysis, we fixed the preprocessing configuration to isolate and evaluate the impact of flood mapping hyperparameters. For instance, Figure \ref{fig:hyperparametersimpact} shows the effect of hyperparameters for the active contour model for a fixed speckle filtering configuration (SAR2SAR without hyperparameters). With this speckle filtering configuration, the sensitivity to hyperparameters is evident for the active contour model, revealing a significant spread in Accuracy and F1-score across different hyperparameter configurations. It highlights the importance of hyperparameter tuning for this method. Thresholding and change detection methods exhibit much lower sensitivity to hyperparameters (given the sampling used in this study). 

\begin{figure*}[ht!]
    \centering
    \includegraphics[width=1\linewidth]{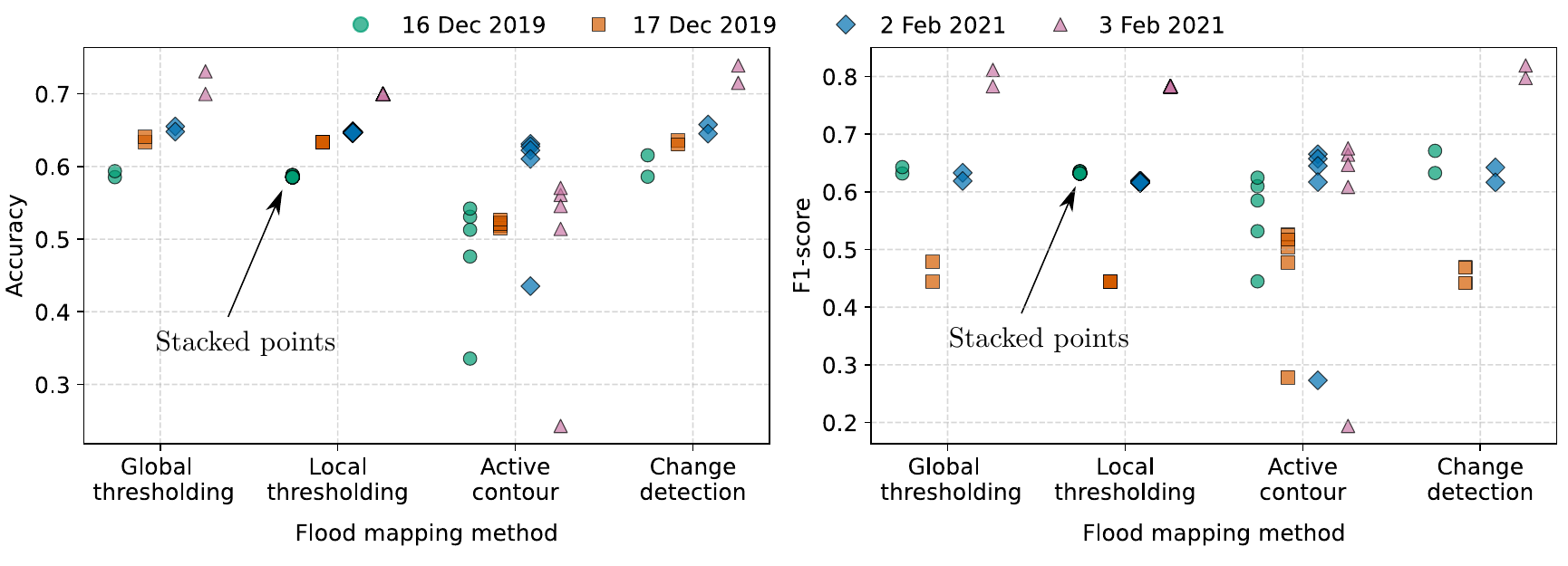}
    \caption{Impact on Accuracy and F1-score of different flood mapping configurations with fixed speckle filtering (SAR2SAR approach) for the four satellite acquisitions.}
    \label{fig:hyperparametersimpact}
\end{figure*}

A more detailed analysis, similar to that in Figure \ref{fig:preprocessingsensitivity}, was conducted to determine the range of variability due to the hyperparameters of flood mapping methods for all 26 speckle filtering configurations (3 configurations for Median and Lee filters, nine configurations for Lee Sigma and Frost filters, one configuration for SAR2SAR and no preprocessing). The results for the acquisition on 2 February 2021 are presented in Figure \ref{fig:hyperparametersensitivity}. 

 \begin{figure*}[ht!]
     \centering
     \includegraphics[width=1\linewidth]{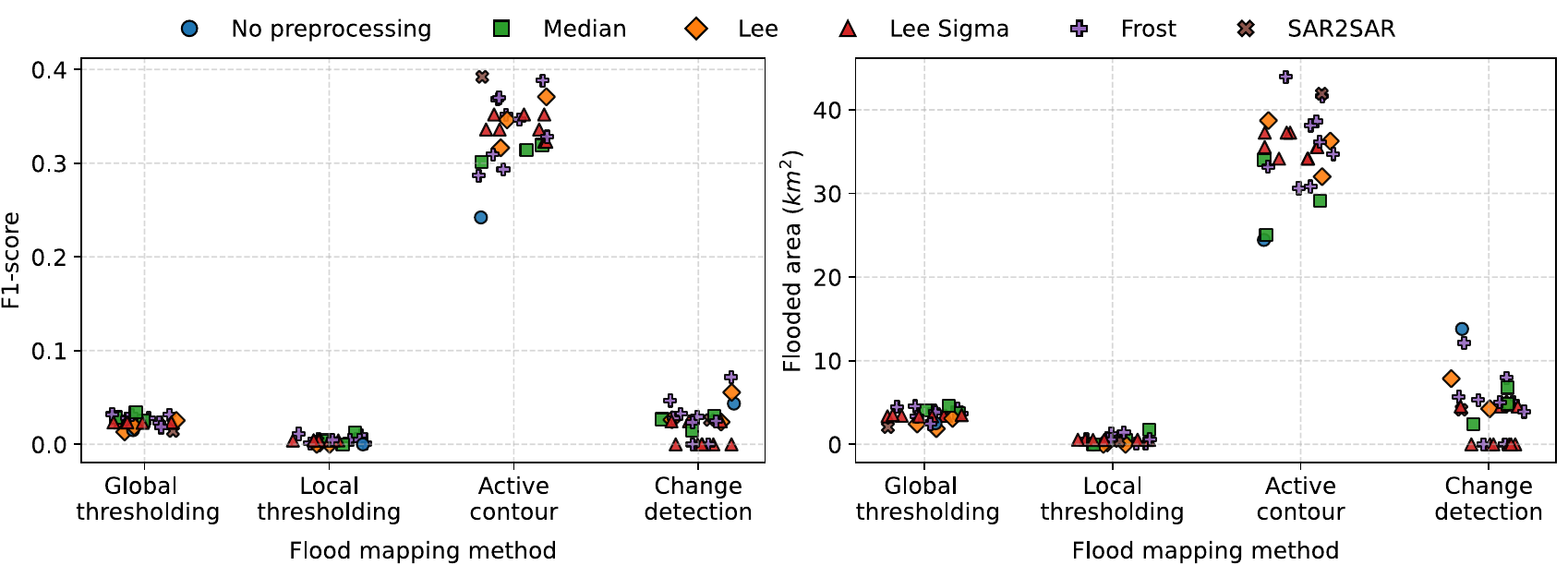}
     \caption{Range of variation (between the minimum and maximum values) of F1-score and flooded area due to flood mapping methods hyperparameters for the 26 speckle filtering configurations for 2 February 2021 acquisition. The points are jittered horizontally for visualization
purposes.}
     \label{fig:hyperparametersensitivity}
 \end{figure*}
 
In this study, the active contour method showed the highest sensitivity to flood mapping hyperparameter variations, with F1-score differences reaching up to 0.34 and differences in flooded surfaces ranging between 30 and 40 km² for all speckle filtering configurations. In contrast, change detection and local thresholding methods demonstrated low sensitivity, with F1-score differences of less than 0.03 and flooded area variations limited to just a few square kilometers. For active contour, most of the variability can be attributed to the flood mapping hyperparameters. While thresholding methods showed limited variability, both active contour and change detection methods displayed greater sensitivity depending on the fixed speckle filtering configuration. For instance, variations in F1-score or flooded area due to hyperparameters for change detection were smaller for Lee Sigma configurations compared to Median filtering configurations.

\subsubsection{Impact of Morphological Operations}Figure~\ref{fig:morphologicaloperations} shows the improvement or degradation in the F1-score for global thresholding, local thresholding, and change detection approaches for one of the acquisitions (2 February 2021) when using morphological operations. The color intensity represents the average gain (or loss) relative to the configurations without morphological post-processing. The active contour and supervised classification approaches are not reported here, as morphological operations had negligible or no effect on the generated flood maps due to the minimal presence of holes or small unconnected flood elements in those flood maps. 

\begin{figure*}[ht!]
    \centering
    \includegraphics[width=0.9\linewidth]{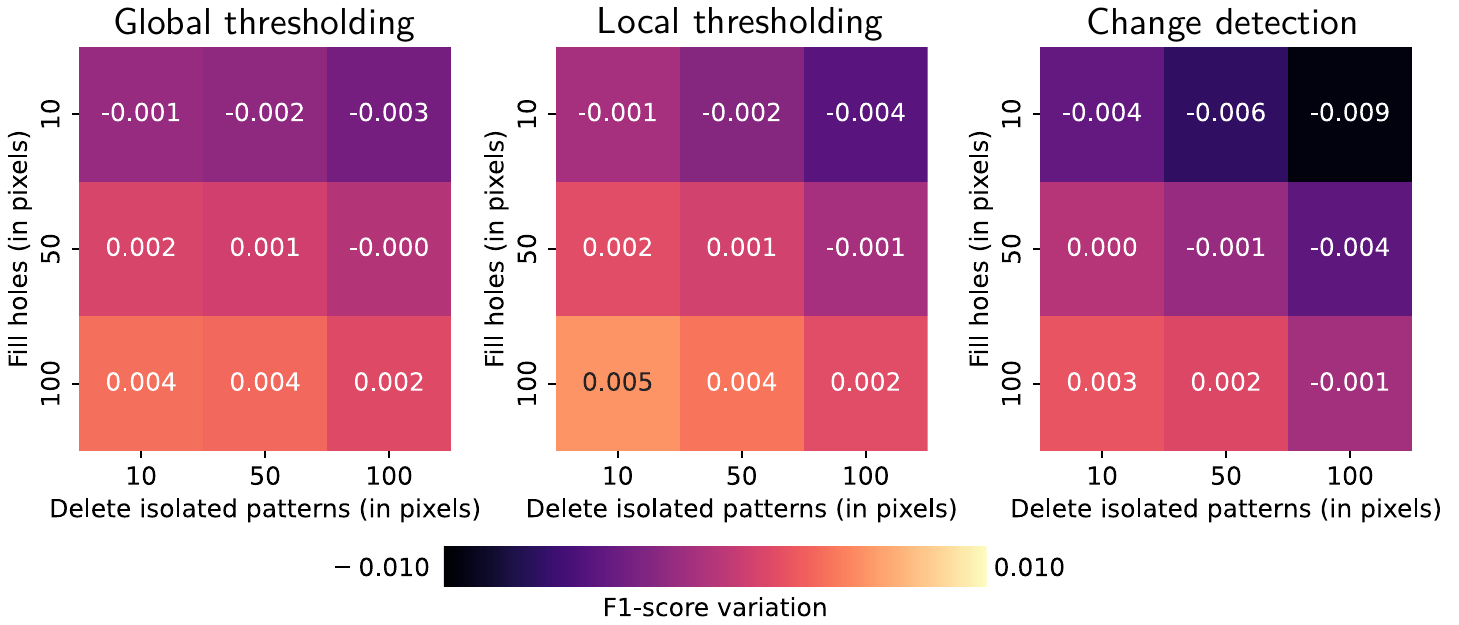}
    \caption{Heatmaps representing the average gain (or loss) in F1-score due to morphological operations relative to configurations without morphological post-processing.}
    \label{fig:morphologicaloperations}
\end{figure*}

We observed that the F1-score increased for the three methods for large hole filling (50 to 100 pixels) and small patch removal (10 to 50 pixels). This is mainly due to eliminating isolated false positives and filling small gaps, improving the spatial coherence of flood maps. However, with the increasing size of patch removal, the performance decreases, suggesting that the small-scale features are well-captured and should not be removed. The F1-score improvements are smaller for change detection since the generated flood maps were closer to the reference than for thresholding before applying morphological operations. While the role of morphological operations was evaluated, the relative improvements in the F1-score metric are relatively low.

Finally, Figure \ref{fig:bestF1} compares the flood maps with the highest F1-scores obtained for each method, including those improved with morphological operations for 2 February 2021. 

\begin{figure*}[ht!]
    \centering
    \includegraphics[width=0.89\linewidth]{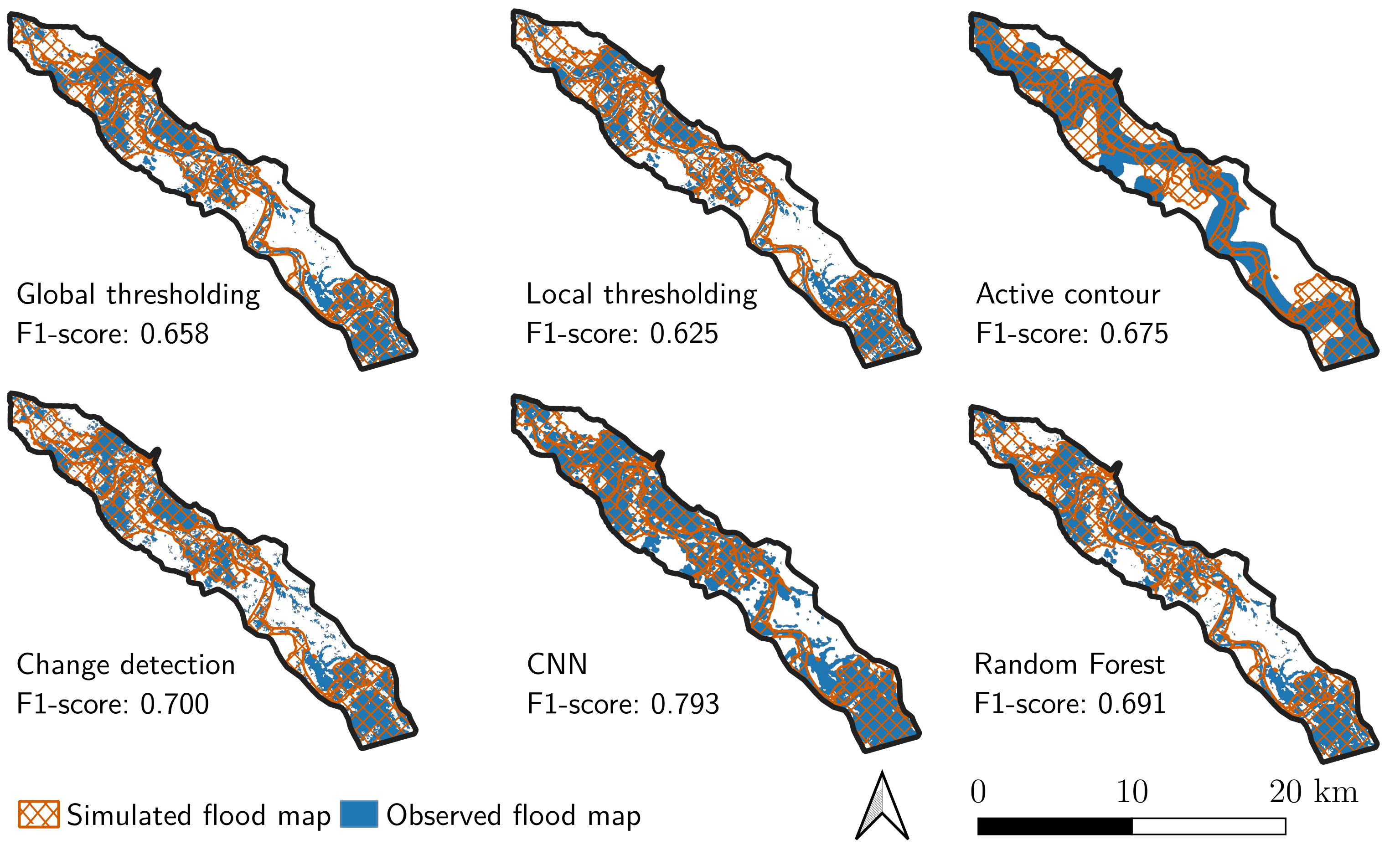}
    \caption{Comparison of the simulated flood maps (orange) with the generated flood maps (blue) for 2 February 2021.}
    \label{fig:bestF1}
\end{figure*}

The CNN-based approach achieved the best performance, with an F1-score of 0.793, followed by the Change Detection and Random Forest methods. The flood map produced by the CNN is the most continuous and closely aligns with the simulated flood map. The active contour model also exhibits a continuous flood extent, but it significantly underestimates inundation in both the upstream and downstream regions. The other methods result in less smoothed flood extent, capturing more localized details and irregularities. While some of these details may correspond to actual observations, their fragmented nature results in poorer alignment with the continuous structure of the simulated flood map.

\FloatBarrier

\section{Water depth estimation}\label{sec:depthestimation} The flood maps extracted from the preprocessed SAR images can be used with other inputs, such as Digital Elevation Models (DEM), to construct water depth fields. In the literature, the approach of \cite{hostache2009water} for water depth estimation has proved effective but requires prior expertise in the flow motions and extensive fieldwork to constrain the method. In this study, we preferred to compare automatic water depth estimation methods on the whole flood map or specific cross sections. Methods such as the Flood Water Depth Estimation Tool (Fw-DET) \citep{cohen2018estimating} or the FLEXTH methodology \citep{betterle2024water} are available to derive water depth fields across the entire domain based on a flood map and DEM. Other methods rely on cross-section analysis by considering that the free surface is flat on a cross-section, thus retrieving the water depth field by knowing the locations of the edges of the flood and topography. We generated water depth fields by applying these methods across a range of hyperparameters, as summarized in Table \ref{table:all_methods}. The water depth estimation workflow follows a structured sequence illustrated in Figure \ref{fig:waterdepthworkflow}:
\begin{enumerate}
\item \textbf{Input:} Flood maps derived from SAR imagery and a Digital Elevation Model (DEM), all projected on the same spatial grid.
\item \textbf{Methods:} Fw-DET, FLEXTH, or cross-section analysis. Each method is evaluated for different hyperparameters.
\item \textbf{Water depth estimation:} Each method determines the water surface elevation and subtracts the terrain elevation from the DEM to generate a water depth field.
\end{enumerate}

\begin{figure*}[ht!]
    \centering
    \includegraphics[width=1\linewidth]{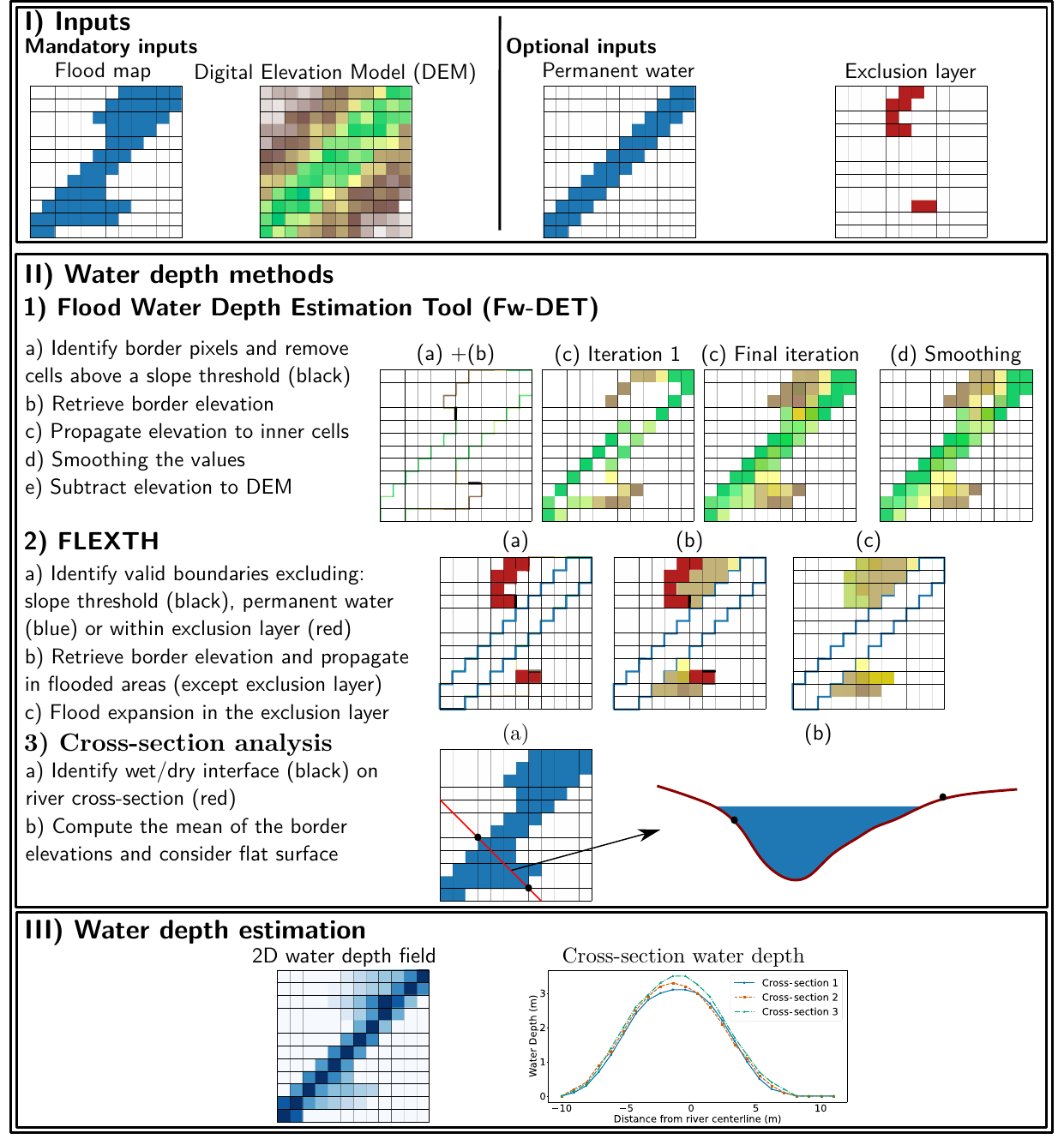}
    \caption{Schematic overview of the water depth estimation methodology.}
    \label{fig:waterdepthworkflow}
\end{figure*}

\subsection{Methods}

\subsubsection{Fw-DET method}
The Flood Water Depth Estimation Tool (Fw-DET) \citep{cohen2018estimating,cohen2019floodwater} quantifies water depth continuously in the domain using a flood extent polygon and a Digital Elevation Model. A schematic representation of the Fw-DET method is presented in panel II)1) of Figure \ref{fig:waterdepthworkflow}. Flood boundaries are derived from the flood map and converted into a line layer, while steep slope cells are filtered out using a slope threshold as a hyperparameter. This line layer is rasterized to align with the DEM grid. Elevation values are extracted for these boundary cells, and each cell within the flood extent polygon is assigned the elevation of the nearest boundary cell under the assumption of a flat water surface. Water depth is computed by subtracting the DEM elevation from the water surface elevation. To mitigate artifacts caused by mismatches between the DEM and the flood extent, a smoothing procedure is applied multiple times using a $3 \times 3$ window.

This study considered the number of smoothing iterations and the slope threshold as model parameters. The latest version of the code developed by \cite{cohen2019floodwater} is available at \url{https://github.com/csdms-contrib/fwdet}. For the present study, the code was adapted to operate without the QGIS interface, using standalone Python scripts to enable batch processing.
\subsubsection{FLEXTH method}
The FLEXTH method \citep{betterle2024water} presents an approach similar to the Fw-DET methodology but introduces improvements to mitigate unrealistic water depth estimates \citep{cohen2019floodwater}. Figure \ref{fig:waterdepthworkflow}, panel II)2), depicts the FLEXTH methodology schematically. The methodology was extended to account for additional inputs, such as exclusion or permanent/seasonal water body masks. The exclusion layer indicates areas where the satellite sensor cannot accurately discriminate flooded areas from dry areas due to water-look-alike conditions (e.g., roads, dikes, vegetated areas). The method expands the flooded area into adjacent no-data regions and estimates the water depths based on the DEM \citep{betterle2024water}. As in Fw-DET, pixels along the flood boundaries with a slope exceeding a user-defined threshold are excluded from the boundary. Water depths within the flooded area are then estimated using a weighted average of the boundary cell elevations based on the closest cells up to a specified maximum number of neighbors. 

The slope threshold and the maximum number of neighbors used in the computation were considered hyperparameters. In \cite{betterle2024water}, they highlighted that these two parameters are the most influential in the FLEXTH method.
The FLEXTH method, initially developed by \cite{betterle2024water}, is available as a Python code at \url{https://code.europa.eu/floods/floods-river/flexth} and was adapted for the needs of our study.

\subsection{Cross-section approach}
The two previous methods estimated the water depths continuously across the domain. An alternative approach is to perform cross-section analysis along the river channel by extracting the surface elevation at the dry/flood interface along predefined cross-sections. The water surface is assumed to be flat for each cross-section, and its elevation is computed as the mean elevation of the identified boundary cells. However, this method is sensitive to errors caused by over-detection of wet/dry interfaces along the cross-sections. To address this issue, an alternative strategy involves considering only the left and right banks of the flood extent \citep{schumann2007high}, excluding other points within the flooded area. This strategy is displayed schematically in panel II)3) of Figure \ref{fig:waterdepthworkflow}. In this study, we tested this second approach. The cross-section approach did not use any hyperparameters in this study.

\subsection{Results}
The methods for estimating water depth fields were applied to the flood maps generated in Section \ref{sec:floodmapping}. For each satellite acquisition, more than 10,000 flood maps were generated. In our implementation, the processing time to generate the water depth field for each flood map was approximately 15 seconds for Fw-DET, 30 seconds to 1 minute for FLEXTH, and only a few seconds for the cross-section approach. All computations were conducted on an Intel® Core™ i7-11850H CPU @ 2.50GHz. 
Thus, generating water depth fields for all hyperparameter combinations (nine configurations of the Fw-DET method, nine configurations of the FLEXTH method, and one for the cross-section approach) would result in an impractically large number of outputs and computation time. Since the generation of water depth fields depends on the input flood contour, our focus is on analyzing the variations of the water depth fields due to variations in the flood contour (due to speckle filtering, flood mapping methods, and hyperparameters used previously). We selected 10 flood maps per flood mapping method to capture a representative range of potential flood map inputs. These were sampled uniformly across the full span of their F1-scores, from the lowest to the highest score, to ensure coverage of both high- and low-quality segmentation results. The Fw-DET, FLEXTH, and cross-section methods were applied for these flood maps. For the cross-section approach, the results were analyzed for two user-defined cross-sections shown in Figure \ref{fig:crossections}. The estimated water depths were compared with the hydrodynamic simulations and watermarks presented in Section \ref{sec:validationdata}. The watermarks were used only for the 16 December 2019 and 3 February 2021 acquisitions (near the flood's peak). For the comparison against simulations, the Root Mean Square Error (RMSE) was computed, comparing pixel-to-pixel values. The rasterized results were projected at the measurements' locations for the watermarks. 

\begin{figure}[ht!]
    \centering
    \includegraphics[width=0.5\linewidth]{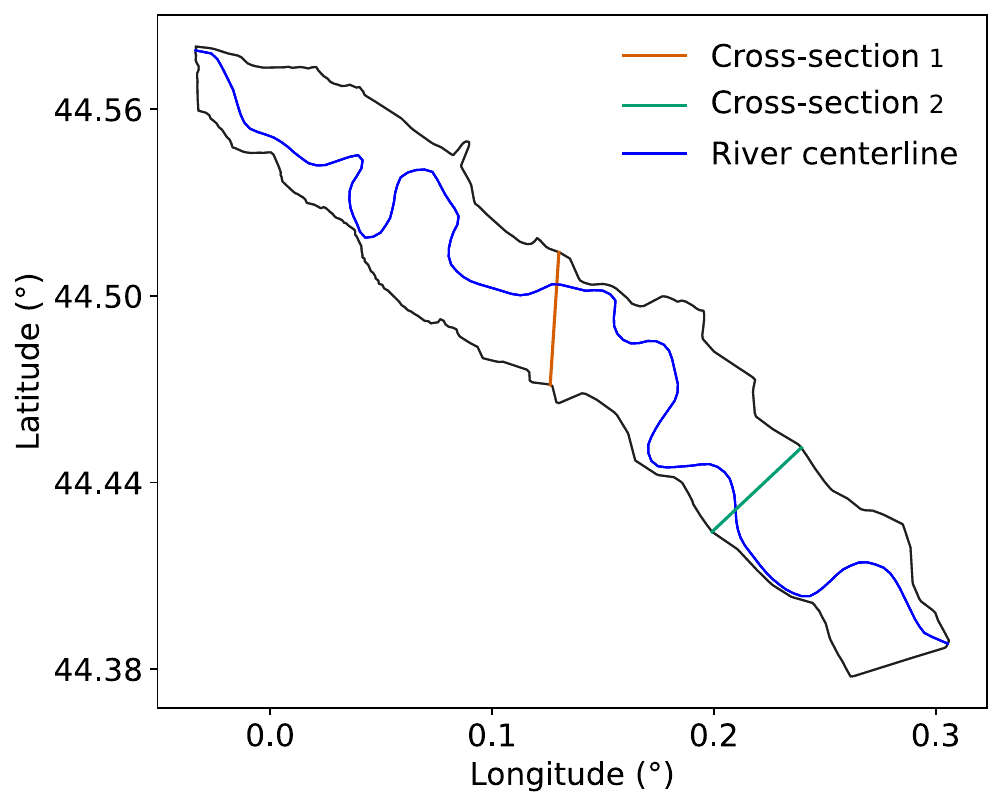}
    \caption{Visualization of the two cross-sections (cross-section 1 in orange and cross-section 2 in green) used in the study.}
    \label{fig:crossections}
\end{figure}

\subsubsection{Water depth estimation methods evaluation}
Figure \ref{fig:waterdepthestimation} shows the RMSE of the estimated water depths compared to hydrodynamic simulations (see Figure \ref{fig:waterdepthestimation}a) and watermarks (see Figure \ref{fig:waterdepthestimation}b) for the FwDET and FLEXTH methods. Each box plot indicates the variability explained by the input flood map or water depth estimation hyperparameters. 

\begin{figure}[ht!]
    \centering
    \includegraphics[width=1\linewidth]{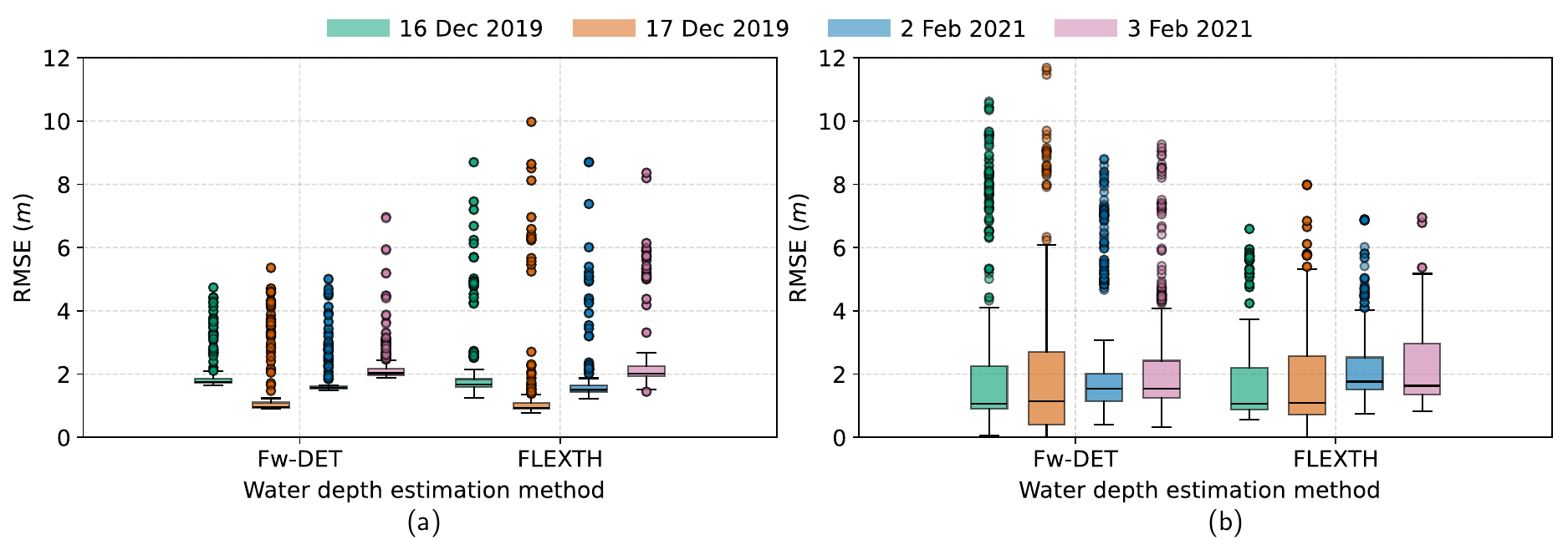}
    \caption{Root Mean Square Error (RMSE) between the estimated water depths and simulated (a) or watermarks (b) for Fw-DET and FLEXTH methods for four satellite images.}
    \label{fig:waterdepthestimation}
\end{figure}

FLEXTH and Fw-DET demonstrate comparable performance in terms of median RMSE. Both methods exhibit a median RMSE ranging from 0.8 to 1.9 m, depending on the satellite image acquisition. The variability of the RMSE is also comparable for both algorithms, with slightly more outliers for the FLEXTH method, which may be caused by the used hyperparameter values. These conclusions are similar either with the simulated water depth fields or watermarks as validation dataset. For the watermarks, the variability is more important. The observed variability and elevated RMSE values can be attributed to the flood map selection from previous processing steps, including high and low F1-score maps relative to the reference. This variability and the influence of water depth estimation method hyperparameters are further analyzed in Section \ref{sec:floodmapsinputs} for the Fw-DET and FLEXTH methods.

Next, Figure \ref{fig:crossectionsanalys} presents the RMSE of water depth estimates on the two selected cross-sections (see Figure \ref{fig:crossections}) for the cross-section approach, along with Fw-DET and FLEXTH methods projected onto these cross-sections.

The Fw-DET and FLEXTH methods exhibited similar RMSE performance for both cross-sections, though FLEXTH showed higher variability. The cross-section approach, while showing similar or slightly higher median RMSE values compared to FLEXTH and Fw-DET on the two cross-sections, exhibited more variability, particularly for the 2 February 2021 image, which was likely related to challenging flood contour detection conditions for that image. The Fw-DET and FLEXTH algorithms propagate the surface elevation from the edges of the flood to the inner flood. The cross-section approach, on the other hand, relies only on the identification of the right and left edges of the flood. Then, the cross-section approach is highly sensitive to identifying the border of the flood on the cross-section, while it is less the case for Fw-DET and FLEXTH, which rely on the whole flood extent borders. The variability of the RMSE for the cross-section approach was entirely due to the variability of the input flood maps, since the cross-section approach was considered without hyperparameters. 

\begin{figure}[ht!]
    \centering
    \includegraphics[width=1\linewidth]{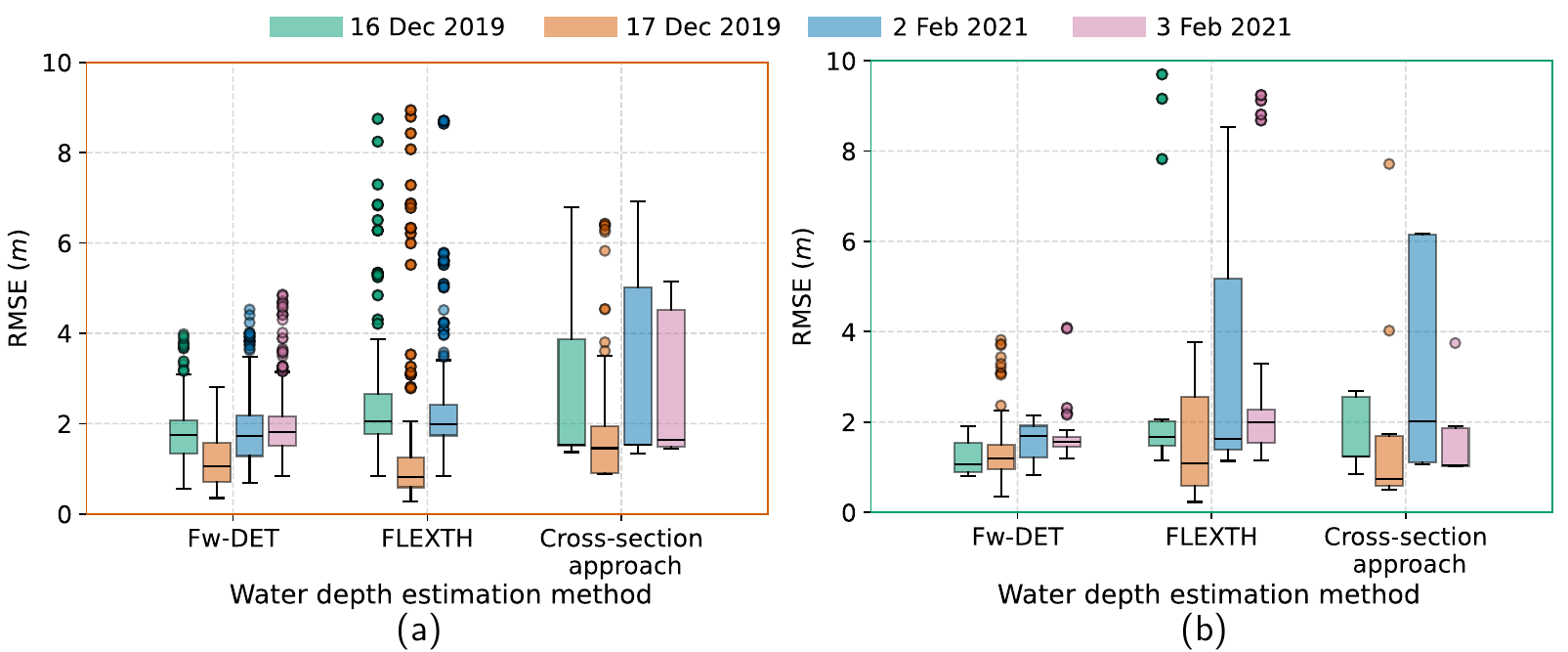}
    \caption{Comparison of Fw-DET, FLEXTH, and cross-section approaches on cross-section 1 (a) and cross-section 2 (b) for four satellite acquisitions against hydrodynamic simulations.}
    \label{fig:crossectionsanalys}
\end{figure}

\subsubsection{Impact of hyperparameters and flood map inputs}\label{sec:floodmapsinputs}
The influence of flood map inputs and method hyperparameters on water depth estimation performance was evaluated for the Fw-DET and FLEXTH methods. Figure \ref{fig:inputfloodmaps} presents the distribution of RMSE values, computed against hydrodynamic simulations across four satellite acquisitions, grouped by flood mapping method used as input to the water depth estimation methodology.

 \begin{figure*}[ht!]
     \centering
     \includegraphics[width=1\linewidth]{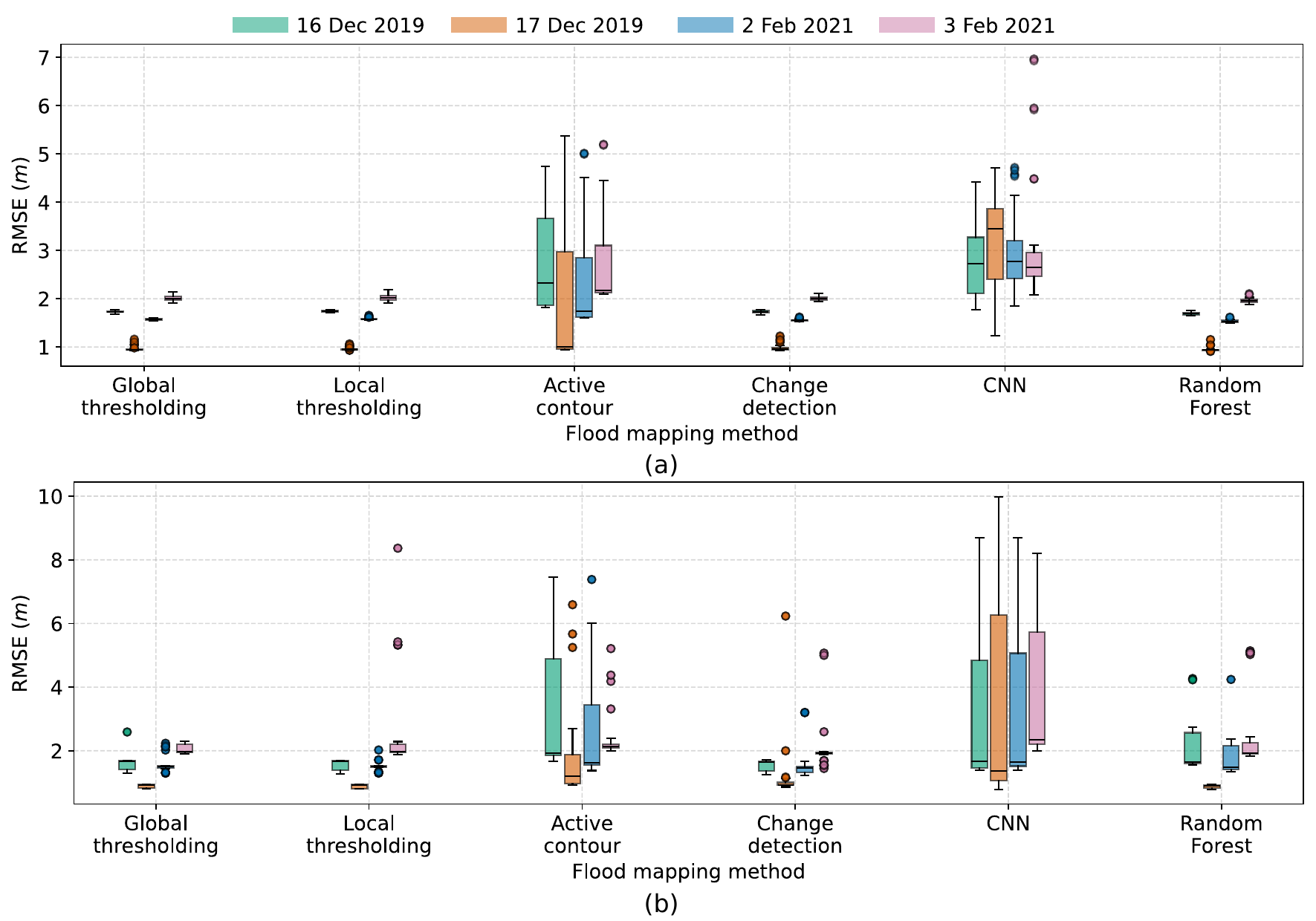}
     \caption{Impact of the flood mapping method used for water depth estimation for the Fw-DET method (a) and FLEXTH method (b) across four satellite acquisitions.}
     \label{fig:inputfloodmaps}
 \end{figure*}
The results indicated that the flood map used as input for the water depth estimation process influenced the estimation accuracy. Flood maps derived from global and local thresholding showed lower RMSE and variability. In contrast, flood maps produced with CNN and active contour lead to greater variability and higher RMSE in water depth estimates. However, the median RMSE values for CNN and active contour flood mapping methods are close to those of the other algorithms, suggesting that they can result in similar estimations with water depth estimation method parameter tuning. The outputs from CNN and active contour are smoother, but their ability to precisely delineate flood boundaries is reduced due to the smoothness. This can result in significant errors in water depth estimates, particularly in areas with steep terrain.

To study the influence of hyperparameters in the FLEXTH and Fw-DET methodologies, we proceeded similarly as in Section \ref{sec:hyperparametersfloodmapping} for flood mapping methods. For each hyperparameter configuration (9 each for FLEXTH and Fw-DET) of water depth estimation methods, we computed the range between the lowest and maximum RMSE for every flood map input. Figure \ref{fig:hyperparameterswaterdepth} presents a statistical analysis of the variability of this range for all the flood map input configurations for the 2 February 2021 acquisition.

\begin{figure}[ht!]
    \centering
    \includegraphics[width=0.5\linewidth]{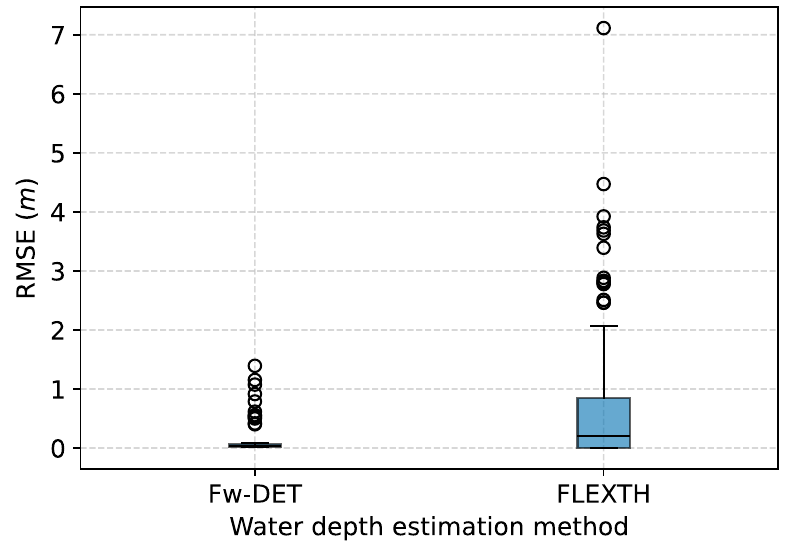}
    \caption{Box plot analysis of the variability induced only by water depth estimation hyperparameters for the 2 February 2021 acquisition.}
    \label{fig:hyperparameterswaterdepth}
\end{figure}

While Fw-DET showed minimal variability in RMSE due to its hyperparameters (except for some outliers beyond the interquartile range), the FLEXTH method displays a much larger interquartile range of RMSE values. In this study, the Fw-DET method demonstrates more consistent performance. In contrast, FLEXTH, while potentially more accurate in ideal cases, was more sensitive to hyperparameter choices and could produce larger errors depending on the configuration used in each test case.

\FloatBarrier

\section{Discussions}\label{sec:discussion}
This study evaluated the influence of preprocessing, flood mapping, and water depth estimation methods using SAR imagery for hydraulic applications. Our results highlighted that each stage, from speckle filtering to method choices and hyperparameter tuning, introduced variability that can propagate through to the final outputs.

\subsection{Limitations of the evaluation process}
The first limitation of the study was the unavailability of ground truth data for flood maps and water depth fields in the study area. As an alternative, hydrodynamic simulations were used as a reference for evaluating performance. These simulations introduce their uncertainties due to discretization, modeled physical processes, or roughness parameterization. Then, some of the observed discrepancies between estimated and reference values may stem from errors in the hydrodynamic model rather than from the methods used to extract the information in this study. Although the availability of a reliable reference would enhance validation, the primary goal of this study was to quantify the variability introduced by preprocessing, method choices, and hyperparameter settings. However, to select the most accurate method, it is essential to use test cases where independent ground truth is available. For instance, \cite{landuyt2018flood} used Copernicus Emergency Management Service flood maps extracted from SAR or optical data, and \cite{li2018automatic} used labeled optical images (in cloud-free conditions) to validate SAR-based flood maps. 

In this study, hyperparameter tuning was not uniformly exhaustive across all methods. For instance, traditional flood mapping methods were evaluated across a range of hyperparameter settings, while supervised classification methods were applied using fixed configurations. This may limit their performance and underestimate their potential under optimal settings or hide their variability due to hidden hyperparameters. Although
we analyzed a range of key hyperparameters, additional influential parameters may be embedded within the methods themselves. Nevertheless, we believe that the primary sources of uncertainty were adequately captured by the parameters examined in this study. Future work could expand the range of hyperparameters considered and apply uncertainty quantification techniques, such as sensitivity analysis, to quantify the influence of each parameter on the outcomes by using a more exhaustive Design of Experiments.

\subsection{Hydraulic analysis and implications for flood monitoring}
The results of this study highlight that the quality of flood maps and water depths estimates is highly sensitive to preprocessing choices, method selection, and hyperparameter tuning. When calibrating models with highly variable outputs, it may propagate errors into the hydrodynamic models, leading to inaccurate parameter estimation. For instance, Figure \ref{fig:floodedarea} highlights that most configurations led to an underestimation of the flooded surface area compared to simulations. First, a fixed Strickler roughness parameter was used across the floodplain and channel subdomains. If the variability due to flood map processing is more important than the changes in simulated flood maps due to the roughness parameter, the calibration could be complicated. Ideally, the variability in the observations should be less than the variability in the simulations.

We evaluated the hydrodynamic model for the 2021 flood event using 2,000 parameter samples under three different Strickler roughness configurations described in Table \ref{table:stricklerevaluation}.

\begin{table}[ht!]
\caption{Three tested Design of Experiments (DoE) for Strickler parameterization for numerical model evaluation. DoE 1 is the baseline with fixed channel values and narrow ranges in floodplains, DoE 2 accounts for wider ranges in floodplains, and DoE 3 also considers a varying Strickler across the three channel subdomains.}
\label{table:stricklerevaluation}
\centering
 \begin{tabular}{l c c c}
 \hline
      Land use & DoE 1& DoE 2& DoE 3\\
      \hline
      Main channel & Fixed & Fixed & $\mathcal{U}(30,50)$ \\
      Waterbodies & $\mathcal{U}(32.5, 37.5)$ & $\mathcal{U}(5, 50)$  & $\mathcal{U}(5, 50)$\\
      
      Fields and meadows without crops & $\mathcal{U}(17.5, 22.5)$  & $\mathcal{U}(5, 50)$ & $\mathcal{U}(5, 50)$\\
      Cultivated fields with low vegetation &$\mathcal{U}(15, 20)$ & $\mathcal{U}(5, 50)$ & $\mathcal{U}(5, 50)$ \\
      Cultivated fields with high vegetation & $\mathcal{U}(10,15)$  & $\mathcal{U}(5, 50)$ & $\mathcal{U}(5, 50)$ \\
      Shrublands and undergrowth areas& $\mathcal{U}(8,12)$ & $\mathcal{U}(5, 50)$ & $\mathcal{U}(5, 50)$ \\
      Areas of low urbanization & $\mathcal{U}(8,10)$  & $\mathcal{U}(5, 50)$ & $\mathcal{U}(5, 50)$\\
      Highly urbanized areas &$\mathcal{U}(5,8)$ & $\mathcal{U}(5, 50)$  & $\mathcal{U}(5, 50)$\\
      \hline
    \end{tabular}
\end{table}

Figure \ref{fig:comparisonDoefloodmaps} compares the 95\% variability range in simulated flooded area for each DoE with the flooded areas obtained using various flood mapping configurations for the February 2, 2021, SAR acquisition. The variability in the simulated flooded area is limited for DoE 1 and was lower than the variability across SAR-derived flood mapping methods. This complicates calibration because uncertainty from SAR processing dominates over that from model parameters. For the second DoE, the larger variation in floodplain roughness resulted in a wider range of flooded areas (approximately 5 km²), which is comparable to the variability observed for each SAR method, except for the active contour method. For DoE 3, there is a large spread in simulated flood extents, which is similar to the spread between methods (for instance, between the change detection median and the CNN median). For this study, only the CNN flood mapping method intersected with the simulation in terms of flooded areas. For the intersection between areas of 71 km² and 75 km², the roughness values in the channel between Mas d'Agenais and La Réole fall at the upper end of the distribution, primarily ranging from 40 to 50 $m^{1/3} \cdot s^{-1}$. In that case, it results in higher roughness values (and thus higher velocities and reduced water depths) than in the previous model calibrations based on stage-gauging stations. The variability introduced by SAR flood mapping methods and hyperparameters exceeds the uncertainty caused by model roughness parameterizations, except for DoE 3. This reinforces the necessity of accounting for uncertainties in flood map generation from SAR images to use them for model calibration.

\begin{figure}[ht!]
    \centering
    \includegraphics[width=1\linewidth]{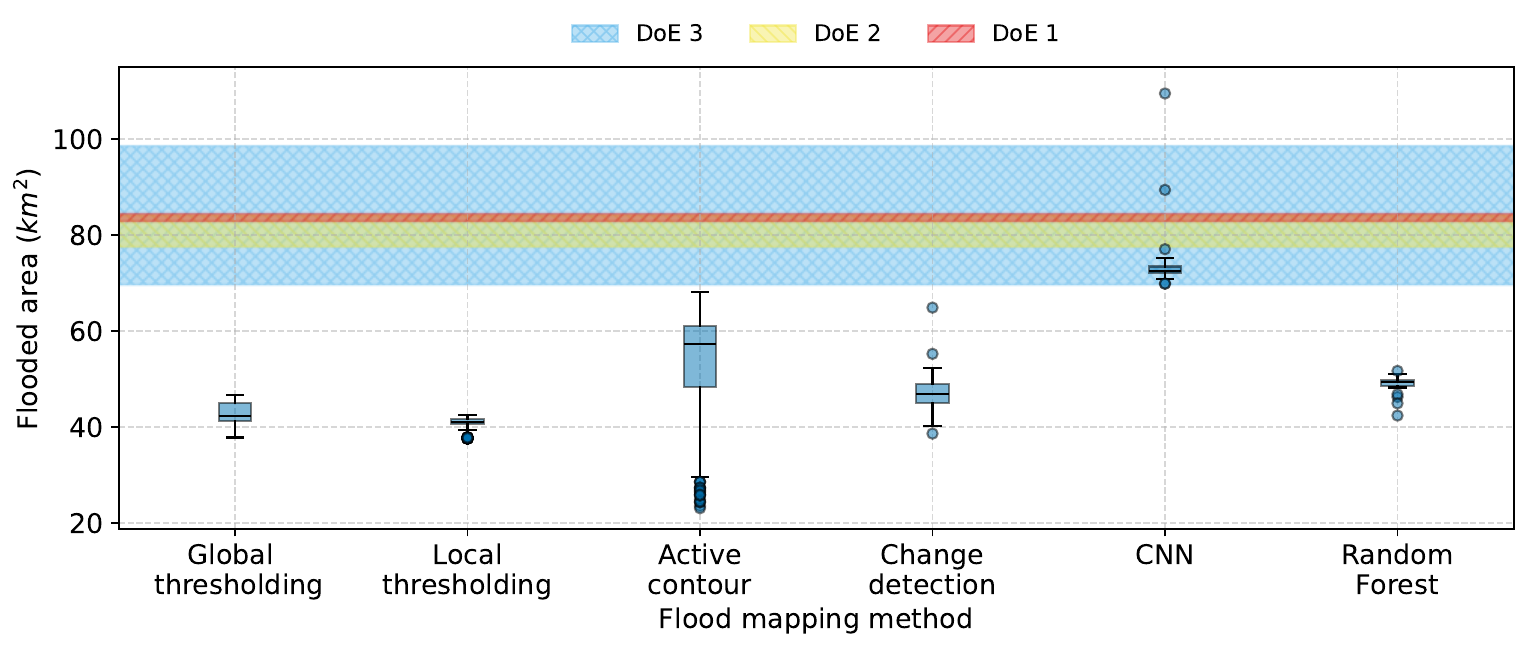}
    \caption{Comparison of flooded surface area variability from hydraulic simulations using three Design of Experiment (DoE 1–3) against SAR-based flood maps. The colored surface for the three DoE indicates the 95\% confidence interval of simulated flooded areas. }
    \label{fig:comparisonDoefloodmaps}
\end{figure}

Water depth estimation methods produced water depth fields with RMSE errors of around 2 m (for the median). These errors are too significant to be ignored in the calibration of this study area. Then, the calibration should be spatially constrained to areas where water depth estimates are more reliable, such as zones with low topographic gradient.

\section{Conclusions}\label{sec:conclusion}
By systematically evaluating the impact of preprocessing, flood mapping, and water depth estimation methods and their hyperparameters, this study aimed to inform best practices for generating flood maps and water depth fields from SAR imagery. It contributes to improving the reliability of SAR-based hydraulic applications by underscoring the importance of assessing the influence of speckle filtering and hyperparameter choices. Rather than relying on a single fixed configuration, the findings highlight the need for a comprehensive evaluation of processing options. These options can significantly impact the extracted flood maps or water depths, ultimately influencing the calibration and validation of hydraulic models. The choice of speckle filtering method was shown to significantly influence the statistics of SAR images, with deep learning methods such as the SAR2SAR approach outperforming the traditional approaches. For flood mapping, supervised classification methods demonstrated the highest median accuracy and F1-score. While less accurate, unsupervised methods such as thresholding and change detection offered reliable results with minimal tuning, making them suitable for rapid deployment. We highlighted the role of the speckle filter method choice and its hyperparameter on the flooded area estimates, with variations in flooded area estimates. The hyperparameters of the flood mapping methods were also influential on the generated flood maps, especially for active contour. Morphological operations provided modest but consistent improvements for specific methods by enhancing the spatial coherence of flood maps.

Three methods for estimating water depth fields were compared, showing that the accuracy of water depth estimation is strongly influenced by both the quality of the input flood maps and the parameterization of the methods. In this study, and for the tested hyperparameters, FLEXTH and Fw-DET exhibited similar performance. For studying the water depth on cross-sections, the cross-section analysis results in similar estimates but with a high variability depending on the cross-sections or input flood maps.

In the future, several limitations of this study could be addressed. Due to the lack of ground truth data for flood map validation, we used hydrodynamic simulations, which include uncertainties due to model errors. Since the study aims to compare output variability caused by preprocessing, method choices, and hyperparameters, the validation dataset is not essential. Furthermore, hyperparameter tuning was not performed for some methods, such as supervised classification, potentially limiting their optimal performance and leaving unexplored uncertainties associated with hidden parameters. Although we analyzed a range of key hyperparameters, additional influential parameters may be embedded within the methods themselves. Nevertheless, we believe that the primary sources of uncertainty were adequately captured by the parameters examined in this study. Future work could expand the range of hyperparameters considered and apply uncertainty quantification techniques, such as sensitivity analysis, to quantify the influence of each parameter on the outcomes.
 
 In future flood studies using SAR imagery for hydraulic applications, the role of preprocessing, method choices, and hyperparameter tuning should be evaluated to account for the errors due to the processing pipeline. We studied only SAR images, but similar methodologies to evaluate the impact of method choices and hyperparameters could be applied to optical satellite images. 
 
\section*{Code and data availibility}
The openTELEMAC software is freely available on a GitLab server with a track of all the developments and a fixed branch for the v8p5 version used in the present study (\url{https://gitlab.pam-retd.fr/otm/telemac-mascaret/-/tree/v8p5r0?ref_type=tags}). The result files of the simulations and their projections on raster grids are provided for easier reproducibility. 
The code and all the data used in this study are available online at
 \url{https://github.com/jtravert/sar-flood-evaluation-framework} with some instructions and Jupyter notebooks to reproduce the methodology on the test case or other study areas and flood events.

\acknowledgments
This work was carried out as part of EDF R\&D's MOISE-2 research project on river and coastal flood hazard assessment, whose support the authors gratefully acknowledge. This work was supported and funded by the French National Association of Research and Technology (ANRT) and EDF R\&D with the
Industrial Conventions for Training through REsearch (CIFRE grant agreement 2022/0972). The authors would like to gratefully acknowledge the open-source community, especially that of the openTELEMAC and of the used Python libraries (scikit-learn, SciPy, rasterio, numpy, matplotlib).
%

\appendix

\end{justify}

\bibliography{template}

\begin{thebibliography}{}

\bibitem[Akiba et~al., 2019]{akiba2019optuna}
Akiba, T., Sano, S., Yanase, T., Ohta, T., and Koyama, M. (2019).
\newblock Optuna: A next-generation hyperparameter optimization framework.
\newblock In {\em Proceedings of the 25th ACM SIGKDD international conference on knowledge discovery \& data mining}, pages 2623--2631.

\bibitem[Ashman et~al., 1994]{ashman1994detecting}
Ashman, K.~M., Bird, C.~M., and Zepf, S.~E. (1994).
\newblock Detecting bimodality in astronomical datasets.
\newblock {\em arXiv preprint astro-ph/9408030}.

\bibitem[Bates, 2012]{bates2012integrating}
Bates, P.~D. (2012).
\newblock Integrating remote sensing data with flood inundation models: how far have we got?
\newblock {\em Hydrological processes}, 26(16):2515--2521.

\bibitem[Bazi et~al., 2005]{bazi2005unsupervised}
Bazi, Y., Bruzzone, L., and Melgani, F. (2005).
\newblock An unsupervised approach based on the generalized gaussian model to automatic change detection in multitemporal sar images.
\newblock {\em IEEE Transactions on Geoscience and Remote Sensing}, 43(4):874--887.

\bibitem[Bentivoglio et~al., 2022]{bentivoglio2022deep}
Bentivoglio, R., Isufi, E., Jonkman, S.~N., and Taormina, R. (2022).
\newblock Deep learning methods for flood mapping: a review of existing applications and future research directions.
\newblock {\em Hydrology and Earth System Sciences}, 26(16):4345--4378.

\bibitem[Besnard and Goutal, 2011]{besnard2011comparaison}
Besnard, A. and Goutal, N. (2011).
\newblock Comparaison de mod{\`e}les 1d {\`a} casiers et 2d pour la mod{\'e}lisation hydraulique d’une plaine d’inondation--cas de la garonne entre tonneins et la r{\'e}ole.
\newblock {\em La Houille Blanche}, (3):42--47.

\bibitem[Betterle and Salamon, 2024]{betterle2024water}
Betterle, A. and Salamon, P. (2024).
\newblock Water depth estimate and flood extent enhancement for satellite-based inundation maps.
\newblock {\em Natural Hazards and Earth System Sciences}, 24(8):2817--2836.

\bibitem[Bhattacharyya, 1943]{bhattacharyya1943measure}
Bhattacharyya, A. (1943).
\newblock On a measure of divergence between two statistical populations defined by their probability distribution.
\newblock {\em Bulletin of the Calcutta Mathematical Society}, 35:99--110.

\bibitem[Bonafilia et~al., 2020]{bonafilia2020sen1floods11}
Bonafilia, D., Tellman, B., Anderson, T., and Issenberg, E. (2020).
\newblock Sen1floods11: A georeferenced dataset to train and test deep learning flood algorithms for sentinel-1.
\newblock In {\em Proceedings of the IEEE/CVF Conference on Computer Vision and Pattern Recognition Workshops}, pages 210--211.

\bibitem[Bovolo and Bruzzone, 2007]{bovolo2007split}
Bovolo, F. and Bruzzone, L. (2007).
\newblock A split-based approach to unsupervised change detection in large-size multitemporal images: Application to tsunami-damage assessment.
\newblock {\em IEEE Transactions on Geoscience and Remote Sensing}, 45(6):1658--1670.

\bibitem[Breiman, 2001]{breiman2001random}
Breiman, L. (2001).
\newblock Random forests.
\newblock {\em Machine learning}, 45(1):5--32.

\bibitem[Brown et~al., 2016]{brown2016progress}
Brown, K.~M., Hambidge, C.~H., and Brownett, J.~M. (2016).
\newblock Progress in operational flood mapping using satellite synthetic aperture radar (sar) and airborne light detection and ranging (lidar) data.
\newblock {\em Progress in Physical Geography}, 40(2):196--214.

\bibitem[Bruniquel and Lopes, 1997]{bruniquel1997multi}
Bruniquel, J. and Lopes, A. (1997).
\newblock Multi-variate optimal speckle reduction in sar imagery.
\newblock {\em International journal of remote sensing}, 18(3):603--627.

\bibitem[Chan and Vese, 2001]{chan2001active}
Chan, T.~F. and Vese, L.~A. (2001).
\newblock Active contours without edges.
\newblock {\em IEEE Transactions on image processing}, 10(2):266--277.

\bibitem[Chini et~al., 2017]{chini2017hierarchical}
Chini, M., Hostache, R., Giustarini, L., and Matgen, P. (2017).
\newblock A hierarchical split-based approach for parametric thresholding of sar images: Flood inundation as a test case.
\newblock {\em IEEE Transactions on Geoscience and Remote Sensing}, 55(12):6975--6988.

\bibitem[Chow et~al., 1988]{chow1988applied}
Chow, V.~T., Maidment, D.~R., and Mays, L.~W. (1988).
\newblock {\em Applied hydrology}.
\newblock McGraw-Hill Book Company.

\bibitem[Clement et~al., 2018]{clement2018multi}
Clement, M.~A., Kilsby, C., and Moore, P. (2018).
\newblock Multi-temporal synthetic aperture radar flood mapping using change detection.
\newblock {\em Journal of Flood Risk Management}, 11(2):152--168.

\bibitem[Cohen et~al., 2018]{cohen2018estimating}
Cohen, S., Brakenridge, G.~R., Kettner, A., Bates, B., Nelson, J., McDonald, R., Huang, Y.-F., Munasinghe, D., and Zhang, J. (2018).
\newblock Estimating floodwater depths from flood inundation maps and topography.
\newblock {\em JAWRA Journal of the American Water Resources Association}, 54(4):847--858.

\bibitem[Cohen et~al., 2019]{cohen2019floodwater}
Cohen, S., Raney, A., Munasinghe, D., Loftis, J.~D., Molthan, A., Bell, J., Rogers, L., Galantowicz, J., Brakenridge, G.~R., Kettner, A.~J., et~al. (2019).
\newblock The floodwater depth estimation tool (fwdet v2. 0) for improved remote sensing analysis of coastal flooding.
\newblock {\em Natural Hazards and Earth System Sciences}, 19(9):2053--2065.

\bibitem[Dalsasso et~al., 2021]{dalsasso2021sar2sar}
Dalsasso, E., Denis, L., and Tupin, F. (2021).
\newblock Sar2sar: A semi-supervised despeckling algorithm for sar images.
\newblock {\em IEEE Journal of Selected Topics in Applied Earth Observations and Remote Sensing}, 14:4321--4329.

\bibitem[Dasgupta et~al., 2021]{dasgupta2021mutual}
Dasgupta, A., Hostache, R., Ramsankaran, R., Schumann, G. J.-P., Grimaldi, S., Pauwels, V.~R., and Walker, J.~P. (2021).
\newblock A mutual information-based likelihood function for particle filter flood extent assimilation.
\newblock {\em Water Resources Research}, 57(2):e2020WR027859.

\bibitem[Deledalle et~al., 2014]{deledalle2014nl}
Deledalle, C.-A., Denis, L., Tupin, F., Reigber, A., and J{\"a}ger, M. (2014).
\newblock Nl-sar: A unified nonlocal framework for resolution-preserving (pol)(in) sar denoising.
\newblock {\em IEEE Transactions on Geoscience and Remote Sensing}, 53(4):2021--2038.

\bibitem[Di~Baldassarre et~al., 2009]{di2009technique}
Di~Baldassarre, G., Schumann, G., and Bates, P.~D. (2009).
\newblock A technique for the calibration of hydraulic models using uncertain satellite observations of flood extent.
\newblock {\em Journal of Hydrology}, 367(3-4):276--282.

\bibitem[El~Garroussi et~al., 2019]{el2019uncertainty}
El~Garroussi, S., de~Lozzo, M., Ricci, S., Lucor, D., Goutal, N., Goeury, C., and Boyaval, S. (2019).
\newblock Uncertainty quantification in a two-dimensional river hydraulic model.
\newblock In {\em International Conference on Uncertainty Quantification in Computational Sciences and Engineering}.

\bibitem[Frost et~al., 1982]{frost1982model}
Frost, V.~S., Stiles, J.~A., Shanmugan, K.~S., and Holtzman, J.~C. (1982).
\newblock A model for radar images and its application to adaptive digital filtering of multiplicative noise.
\newblock {\em IEEE Transactions on pattern analysis and machine intelligence}, (2):157--166.

\bibitem[Giustarini et~al., 2012]{giustarini2012change}
Giustarini, L., Hostache, R., Matgen, P., Schumann, G. J.-P., Bates, P.~D., and Mason, D.~C. (2012).
\newblock A change detection approach to flood mapping in urban areas using terrasar-x.
\newblock {\em IEEE transactions on Geoscience and Remote Sensing}, 51(4):2417--2430.

\bibitem[Giustarini et~al., 2011]{giustarini2011assimilating}
Giustarini, L., Matgen, P., Hostache, R., Montanari, M., Plaza, D., Pauwels, V., De~Lannoy, G., De~Keyser, R., Pfister, L., Hoffmann, L., et~al. (2011).
\newblock Assimilating sar-derived water level data into a hydraulic model: a case study.
\newblock {\em Hydrology and Earth System Sciences}, 15(7):2349--2365.

\bibitem[Goodman, 1976]{goodman1976some}
Goodman, J.~W. (1976).
\newblock Some fundamental properties of speckle.
\newblock {\em JOSA}, 66(11):1145--1150.

\bibitem[Grimaldi et~al., 2016]{grimaldi2016remote}
Grimaldi, S., Li, Y., Pauwels, V., and Walker, J.~P. (2016).
\newblock Remote sensing-derived water extent and level to constrain hydraulic flood forecasting models: Opportunities and challenges.
\newblock {\em Surveys in Geophysics}, 37(5):977--1034.

\bibitem[Henry et~al., 2006]{henry2006envisat}
Henry, J.-B., Chastanet, P., Fellah, K., and Desnos, Y.-L. (2006).
\newblock Envisat multi-polarized asar data for flood mapping.
\newblock {\em International Journal of Remote Sensing}, 27(10):1921--1929.

\bibitem[Hervouet, 2007]{hervouet2007hydrodynamics}
Hervouet, J.-M. (2007).
\newblock {\em Hydrodynamics of free surface flows: modelling with the finite element method}.
\newblock John Wiley \& Sons.

\bibitem[Horritt, 1999]{horritt1999statistical}
Horritt, M. (1999).
\newblock A statistical active contour model for sar image segmentation.
\newblock {\em Image and Vision Computing}, 17(3-4):213--224.

\bibitem[Hostache et~al., 2010]{hostache2010assimilation}
Hostache, R., Lai, X., Monnier, J., and Puech, C. (2010).
\newblock Assimilation of spatially distributed water levels into a shallow-water flood model. part ii: Use of a remote sensing image of mosel river.
\newblock {\em Journal of hydrology}, 390(3-4):257--268.

\bibitem[Hostache et~al., 2009]{hostache2009water}
Hostache, R., Matgen, P., Schumann, G., Puech, C., Hoffmann, L., and Pfister, L. (2009).
\newblock Water level estimation and reduction of hydraulic model calibration uncertainties using satellite sar images of floods.
\newblock {\em IEEE Transactions on Geoscience and Remote Sensing}, 47(2):431--441.

\bibitem[Hostache et~al., 2012]{hostache2012change}
Hostache, R., Matgen, P., and Wagner, W. (2012).
\newblock Change detection approaches for flood extent mapping: How to select the most adequate reference image from online archives?
\newblock {\em International journal of applied earth observation and geoinformation}, 19:205--213.

\bibitem[Hunter, 2005]{hunter2005development}
Hunter, N.~M. (2005).
\newblock {\em Development and assessment of dynamic storage cell codes for flood inundation modelling}.
\newblock PhD thesis, University of Bristol.

\bibitem[Kittler and Illingworth, 1986]{kittler1986minimum}
Kittler, J. and Illingworth, J. (1986).
\newblock Minimum error thresholding.
\newblock {\em Pattern recognition}, 19(1):41--47.

\bibitem[Lai et~al., 2014]{lai2014variational}
Lai, X., Liang, Q., Yesou, H., and Daillet, S. (2014).
\newblock Variational assimilation of remotely sensed flood extents using a 2-d flood model.
\newblock {\em Hydrology and Earth System Sciences}, 18(11):4325--4339.

\bibitem[Landuyt et~al., 2018]{landuyt2018flood}
Landuyt, L., Van~Wesemael, A., Schumann, G. J.-P., Hostache, R., Verhoest, N.~E., and Van~Coillie, F.~M. (2018).
\newblock Flood mapping based on synthetic aperture radar: An assessment of established approaches.
\newblock {\em IEEE Transactions on Geoscience and Remote Sensing}, 57(2):722--739.

\bibitem[LeCun et~al., 2015]{lecun2015deep}
LeCun, Y., Bengio, Y., and Hinton, G. (2015).
\newblock Deep learning.
\newblock {\em nature}, 521(7553):436--444.

\bibitem[Lee, 1980]{lee1980digital}
Lee, J.-S. (1980).
\newblock Digital image enhancement and noise filtering by use of local statistics.
\newblock {\em IEEE transactions on pattern analysis and machine intelligence}, (2):165--168.

\bibitem[Lee, 1983]{lee1983digital}
Lee, J.-S. (1983).
\newblock Digital image smoothing and the sigma filter.
\newblock {\em Computer vision, graphics, and image processing}, 24(2):255--269.

\bibitem[Lee et~al., 1994]{lee1994speckle}
Lee, J.-S., Jurkevich, L., Dewaele, P., Wambacq, P., and Oosterlinck, A. (1994).
\newblock Speckle filtering of synthetic aperture radar images: A review.
\newblock {\em Remote sensing reviews}, 8(4):313--340.

\bibitem[Lee and Pottier, 2017]{lee2017polarimetric}
Lee, J.-S. and Pottier, E. (2017).
\newblock {\em Polarimetric radar imaging: from basics to applications}.
\newblock CRC press.

\bibitem[Lee et~al., 2008]{lee2008improved}
Lee, J.-S., Wen, J.-H., Ainsworth, T.~L., Chen, K.-S., and Chen, A.~J. (2008).
\newblock Improved sigma filter for speckle filtering of sar imagery.
\newblock {\em IEEE Transactions on Geoscience and Remote Sensing}, 47(1):202--213.

\bibitem[Li et~al., 2018]{li2018automatic}
Li, Y., Martinis, S., Plank, S., and Ludwig, R. (2018).
\newblock An automatic change detection approach for rapid flood mapping in sentinel-1 sar data.
\newblock {\em International journal of applied earth observation and geoinformation}, 73:123--135.

\bibitem[Manning et~al., 1890]{manning1890flow}
Manning, R., Griffith, J.~P., Pigot, T., and Vernon-Harcourt, L.~F. (1890).
\newblock {\em On the flow of water in open channels and pipes}.

\bibitem[Martinis, 2010]{martinis2010automatic}
Martinis, S. (2010).
\newblock {\em Automatic near real-time flood detection in high resolution X-band synthetic aperture radar satellite data using context-based classification on irregular graphs}.
\newblock PhD thesis, lmu.

\bibitem[Martinis et~al., 2022]{martinis2022towards}
Martinis, S., Groth, S., Wieland, M., Knopp, L., and R{\"a}ttich, M. (2022).
\newblock Towards a global seasonal and permanent reference water product from sentinel-1/2 data for improved flood mapping.
\newblock {\em Remote Sensing of Environment}, 278:113077.

\bibitem[Martinis et~al., 2015a]{martinis2015fully}
Martinis, S., Kersten, J., and Twele, A. (2015a).
\newblock A fully automated terrasar-x based flood service.
\newblock {\em ISPRS Journal of Photogrammetry and Remote Sensing}, 104:203--212.

\bibitem[Martinis et~al., 2015b]{martinis2015comparing}
Martinis, S., Kuenzer, C., Wendleder, A., Huth, J., Twele, A., Roth, A., and Dech, S. (2015b).
\newblock Comparing four operational sar-based water and flood detection approaches.
\newblock {\em International Journal of Remote Sensing}, 36(13):3519--3543.

\bibitem[Martinis et~al., 2009]{martinis2009towards}
Martinis, S., Twele, A., and Voigt, S. (2009).
\newblock Towards operational near real-time flood detection using a split-based automatic thresholding procedure on high resolution terrasar-x data.
\newblock {\em Natural Hazards and Earth System Sciences}, 9(2):303--314.

\bibitem[Mason et~al., 2012]{mason2012automatic}
Mason, D., Schumann, G.-P., Neal, J., Garcia-Pintado, J., and Bates, P. (2012).
\newblock Automatic near real-time selection of flood water levels from high resolution synthetic aperture radar images for assimilation into hydraulic models: A case study.
\newblock {\em Remote Sensing of Environment}, 124:705--716.

\bibitem[Mason et~al., 2009]{mason2009flood}
Mason, D.~C., Speck, R., Devereux, B., Schumann, G. J.-P., Neal, J.~C., and Bates, P.~D. (2009).
\newblock Flood detection in urban areas using terrasar-x.
\newblock {\em IEEE Transactions on Geoscience and Remote Sensing}, 48(2):882--894.

\bibitem[Mateo-Garcia et~al., 2021]{mateo2021towards}
Mateo-Garcia, G., Veitch-Michaelis, J., Smith, L., Oprea, S.~V., Schumann, G., Gal, Y., Baydin, A.~G., and Backes, D. (2021).
\newblock Towards global flood mapping onboard low cost satellites with machine learning.
\newblock {\em Scientific reports}, 11(1):1--12.

\bibitem[Montanari et~al., 2009]{montanari2009calibration}
Montanari, M., Hostache, R., Matgen, P., Schumann, G., Pfister, L., and Hoffmann, L. (2009).
\newblock Calibration and sequential updating of a coupled hydrologic-hydraulic model using remote sensing-derived water stages.
\newblock {\em Hydrology and Earth System Sciences}, 13(3):367--380.

\bibitem[Morvan et~al., 2008]{morvan2008concept}
Morvan, H., Knight, D., Wright, N., Tang, X., and Crossley, A. (2008).
\newblock The concept of roughness in fluvial hydraulics and its formulation in 1d, 2d and 3d numerical simulation models.
\newblock {\em Journal of Hydraulic Research}, 46(2):191--208.

\bibitem[Mumford and Shah, 1989]{mumford1989optimal}
Mumford, D.~B. and Shah, J. (1989).
\newblock Optimal approximations by piecewise smooth functions and associated variational problems.
\newblock {\em Communications on pure and applied mathematics}.

\bibitem[Nguyen et~al., 2022]{nguyen2022dual}
Nguyen, T.~H., Ricci, S., Piacentini, A., Fatras, C., Kettig, P., Blanchet, G., Pe{\~n}a~Luque, S., and Baillarin, S. (2022).
\newblock Dual state-parameter assimilation of sar-derived wet surface ratio for improving fluvial flood reanalysis.
\newblock {\em Preprint}.

\bibitem[Oberstadler et~al., 1997]{oberstadler1997assessment}
Oberstadler, R., H{\"o}nsch, H., and Huth, D. (1997).
\newblock Assessment of the mapping capabilities of ers-1 sar data for flood mapping: a case study in germany.
\newblock {\em Hydrological processes}, 11(10):1415--1425.

\bibitem[Otsu, 1979]{otsu1979threshold}
Otsu, N. (1979).
\newblock A threshold selection method from gray-level histograms.
\newblock {\em IEEE transactions on systems, man, and cybernetics}, 9(1):62--66.

\bibitem[Pulvirenti et~al., 2013]{pulvirenti2013discrimination}
Pulvirenti, L., Marzano, F.~S., Pierdicca, N., Mori, S., and Chini, M. (2013).
\newblock Discrimination of water surfaces, heavy rainfall, and wet snow using cosmo-skymed observations of severe weather events.
\newblock {\em IEEE transactions on geoscience and remote sensing}, 52(2):858--869.

\bibitem[Ronneberger et~al., 2015]{ronneberger2015u}
Ronneberger, O., Fischer, P., and Brox, T. (2015).
\newblock U-net: Convolutional networks for biomedical image segmentation.
\newblock In {\em Medical image computing and computer-assisted intervention--MICCAI 2015: 18th international conference, Munich, Germany, October 5-9, 2015, proceedings, part III 18}, pages 234--241. Springer.

\bibitem[Schumann et~al., 2010]{schumann2010near}
Schumann, G., Di~Baldassarre, G., Alsdorf, D., and Bates, P. (2010).
\newblock Near real-time flood wave approximation on large rivers from space: Application to the river po, italy.
\newblock {\em Water Resources Research}, 46(5).

\bibitem[Schumann et~al., 2007]{schumann2007high}
Schumann, G., Hostache, R., Puech, C., Hoffmann, L., Matgen, P., Pappenberger, F., and Pfister, L. (2007).
\newblock High-resolution 3-d flood information from radar imagery for flood hazard management.
\newblock {\em IEEE transactions on geoscience and remote sensing}, 45(6):1715--1725.

\bibitem[Schumann et~al., 2008]{schumann2008estimating}
Schumann, G., Pappenberger, F., and Matgen, P. (2008).
\newblock Estimating uncertainty associated with water stages from a single sar image.
\newblock {\em Advances in water resources}, 31(8):1038--1047.

\bibitem[Sezgin and Sankur, 2004]{sezgin2004survey}
Sezgin, M. and Sankur, B.~l. (2004).
\newblock Survey over image thresholding techniques and quantitative performance evaluation.
\newblock {\em Journal of Electronic imaging}, 13(1):146--168.

\bibitem[Tarpanelli and Benveniste, 2019]{tarpanelli2019potential}
Tarpanelli, A. and Benveniste, J. (2019).
\newblock On the potential of altimetry and optical sensors for monitoring and forecasting river discharge and extreme flood events.
\newblock In {\em Extreme Hydroclimatic Events and Multivariate Hazards in a Changing Environment}, pages 267--287. Elsevier.

\bibitem[Travert et~al., 2025]{travert2024}
Travert, J.-P., Boyaval, S., Goeury, C., Bacchi, V., and Zaoui, F. (2025).
\newblock Evaluation of performance measures for comparing flood models with satellite observations.
\newblock {\em Water Resources Research}.

\bibitem[Twele et~al., 2016]{twele2016sentinel}
Twele, A., Cao, W., Plank, S., and Martinis, S. (2016).
\newblock Sentinel-1-based flood mapping: a fully automated processing chain.
\newblock {\em International Journal of Remote Sensing}, 37(13):2990--3004.

\bibitem[Zhao et~al., 2020]{zhao2020design}
Zhao, G., Bates, P., Neal, J., and Pang, B. (2020).
\newblock Design flood estimation for global river networks based on machine learning models.
\newblock {\em Hydrology and Earth System Sciences Discussions}, 2020:1--25.

\end{thebibliography}
\bibliographystyle{apalike}
\end{document}